\documentclass[lettersize,journal]{IEEEtran}
\usepackage{amsmath,amsfonts}
\usepackage{algorithmic}
\usepackage{array}
\usepackage{textcomp}
\usepackage{stfloats}
\usepackage{url}
\usepackage{verbatim}
\usepackage{graphicx}
\def\BibTeX{{\rm B\kern-.05em{\sc i\kern-.025em b}\kern-.08em
    T\kern-.1667em\lower.7ex\hbox{E}\kern-.125emX}}
\usepackage{balance}
\usepackage{graphics}
\usepackage{subfigure}
\usepackage{booktabs}
\usepackage{multirow}
\usepackage{cite}
\usepackage{tikz,xcolor,hyperref}
\usepackage{algorithm}
\usepackage{fontawesome}

\definecolor{lime}{HTML}{A6CE39}
\DeclareRobustCommand{\orcidicon}{%
	\begin{tikzpicture}
	\draw[lime, fill=lime] (0,0) 
	circle [radius=0.16] 
	node[white] {{\fontfamily{qag}\selectfont \tiny ID}};
	\draw[white, fill=white] (-0.0625,0.095) 
	circle [radius=0.007];
	\end{tikzpicture}
	\hspace{-2mm}
}

\foreach \x in {A, ..., Z}{%
	\expandafter\xdef\csname orcid\x\endcsname{\noexpand\href{https://orcid.org/\csname orcidauthor\x\endcsname}{\noexpand\orcidicon}}
}

\begin{document}
\title{FE-Fusion-VPR: Attention-based Multi-Scale Network Architecture for Visual Place Recognition by Fusing Frames and Events}
\author{Kuanxu Hou\orcidA{}, Delei Kong\orcidB{}, Junjie Jiang\orcidC{}, Hao Zhuang\orcidD{}, Xinjie Huang\orcidE{} and Zheng Fang\orcidF{}, \emph{Member, IEEE}
\thanks{
Manuscript created March 1, 2022. This work was supported by National Natural Science Foundation of China (62073066, U20A20197), Intel Neuromorphic Research Community (INRC) Grant Award (RV2.137.Fang), Science and Technology on Near-Surface Detection Laboratory (6142414200208), the Fundamental Research Funds for the Central Universities (N2226001), and Aeronautical Science Foundation of China (No.201941050001). \emph{(Corresponding author: Zheng Fang.)}

Kuanxu Hou, Junjie Jiang, Xinjie Huang and Zheng Fang are with Faculty of Robot Science and Engineering, Northeastern University, Shenyang, China (e-mail: 2001995@stu.neu.edu.cn, 2001998@stu.neu.edu.cn, 2101979@stu.neu.edu.cn, fangzheng@mail.neu.edu.cn).

Delei Kong and Hao Zhuang are with College of Information Science and Engineering, Northeastern University, Shenyang, China (e-mail: kong.delei.neu@gmail.com, 2100922@stu.neu.edu.cn).
}}

\markboth{IEEE Robotics and Automation Letters, Vol. ?, No. ?, November 2022}%
{How to Use the IEEEtran \LaTeX \ Templates}

\maketitle

\begin{abstract}
Traditional visual place recognition (VPR), usually using standard cameras, is easy to fail due to glare or high-speed motion. By contrast, event cameras have the advantages of low latency, high temporal resolution, and high dynamic range, which can deal with the above issues. Nevertheless, event cameras are prone to failure in weakly textured or motionless scenes, while standard cameras can still provide appearance information in this case. Thus, exploiting the complementarity of standard cameras and event cameras can effectively improve the performance of VPR algorithms. In the paper, we propose FE-Fusion-VPR, an attention-based multi-scale network architecture for VPR by fusing frames and events. First, the intensity frame and event volume are fed into the two-stream feature extraction network for shallow feature fusion. Next, the three-scale features are obtained through the multi-scale fusion network and aggregated into three sub-descriptors using the VLAD layer. Finally, the weight of each sub-descriptor is learned through the descriptor re-weighting network to obtain the final refined descriptor. Experimental results show that on the Brisbane-Event-VPR and DDD20 datasets, the Recall@1 of our FE-Fusion-VPR is 29.26\% and 33.59\% higher than Event-VPR and Ensemble-EventVPR, and is 7.00\% and 14.15\% higher than MultiRes-NetVLAD and NetVLAD. To our knowledge, this is the first end-to-end network that goes beyond the existing event-based and frame-based SOTA methods to fuse frame and events directly for VPR.

\end{abstract}

\begin{IEEEkeywords}
Visual place recognition, event camera, attention mechanism, multi-scale network, visual sensor fusion.
\end{IEEEkeywords}

\section{Introduction}
\label{sec:introduction}

\begin{figure}[htbp]
\vspace{0mm}
\centering
\includegraphics[width=\columnwidth]{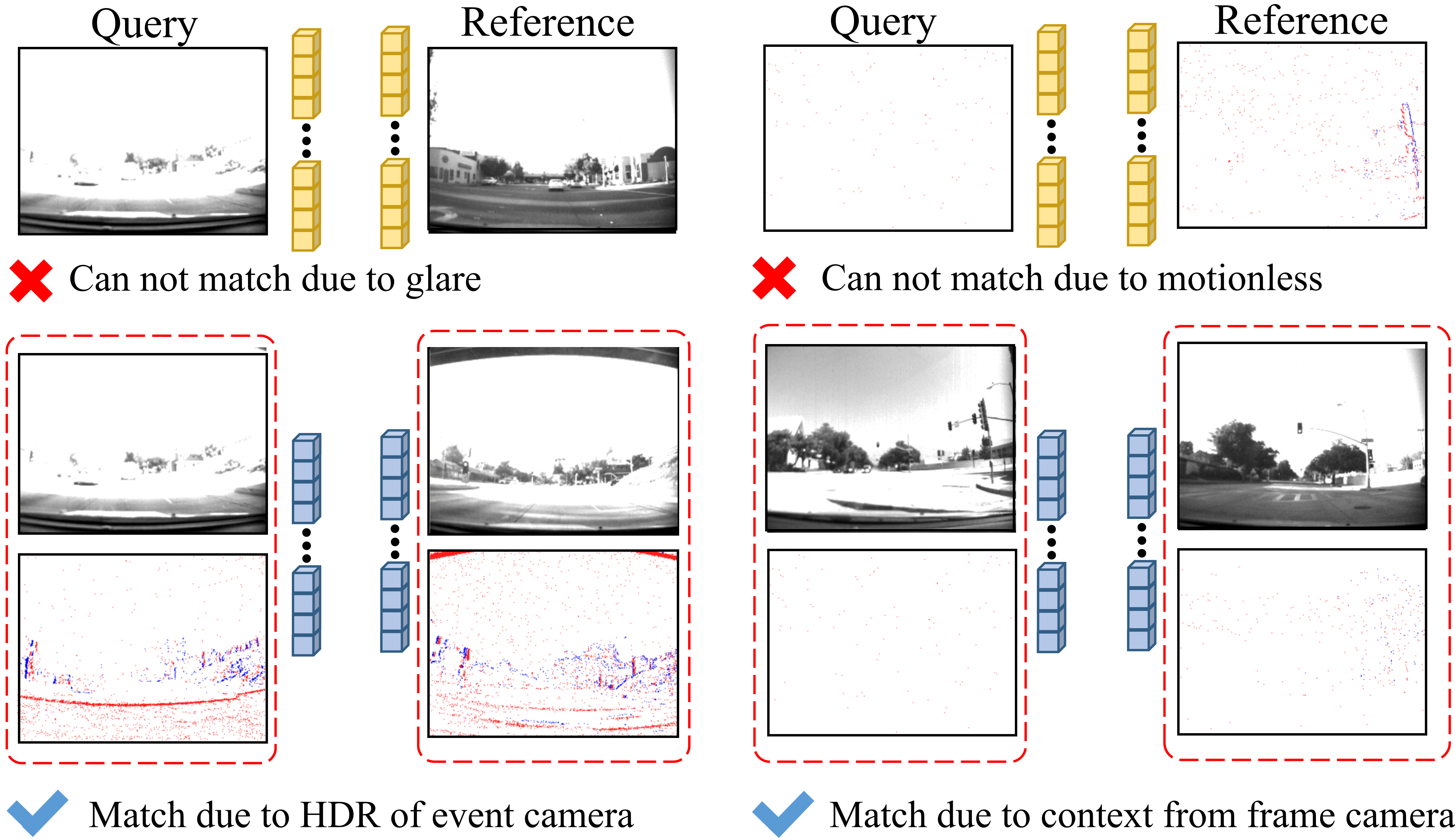}
\caption{Illustration of advantages of the proposed FE-Fusion-VPR. As can be seen, both frames and events have challenging scenarios that are difficult to deal with alone, so combining frames and events can effectively deal with complex scenarios and improve the performance of VPR pipeline.}
\label{fig:1}
\vspace{0mm}
\end{figure}

\IEEEPARstart{V}{isual} place recognition (VPR) \cite{lowry2015visual, zhang2021visual, masone2021survey, garg2021your} is a vital sub-problem in the autonomous navigation of mobile robots, which has attracted the attention of many researchers in recent years. VPR aims to help a robot determine whether it locates in a previously visited place. Specifically, there is an existing database about the environment, which stores visual data (such as frames) for various places in the environment. Now given a query data, we expect to obtain its location information by finding database data captured at the same (or close) location as the query data. In a word, VPR can assist mobile robots or autonomous unmanned systems in localization and loop closure detection (LCD) in GPS-denied environments.

Currently, since standard frame cameras could provide rich appearance information of everyday scenes, existing frame-based VPR methods could achieve good performance in those scenarios. However, standard cameras suffer from low frame rate, motion blur, and sensitivity to illumination changes, which makes traditional VPR methods difficult to handle challenging scenes (e.g., high speed and high dynamic range). Event cameras \cite{delbruck2010activity, gallego2020event, chen2020event, wu2021novel}, which are neuromorphic vision sensors, record microsecond-level pixel-wise brightness changes and offer significant advantages (e.g., low latency, rich motion information and high dynamic range). Nevertheless, event cameras lack appearance (texture) information in some cases (e.g., still or low-speed scenes). Therefore, the above two kinds of vision sensors are complementary. As shown in Fig. \ref{fig:1}, we present two representative examples (glare and motionless) respectively, and these challenging situations can be solved by visual sensor fusion.

Inspired by their complementarity, we consider combining standard frame cameras and event cameras to lift the limitations faced by single vision sensors in large-scale place recognition problems, thereby improving the performance of the VPR pipeline in challenging scenarios. Recently, there have been some works trying to combine frames and events for some machine vision tasks and achieve excellent results \cite{gehrig2018asynchronous,jiang2019mixed,hu2020ddd20,gehrig2021combining,pan2020single,lee2022fusion,vidal2018ultimate,jung2020constrained}. However, combining frames and events for VPR is not very straightforward. We need to deal with several challenges: (1) How to fuse raw frame with event data? (2) How to extract multi-scale features? (3) How to fuse multiple features into a unified descriptor? Addressing the above challenges, in this paper, we propose a robust VPR method by fusing frames and events (FE-Fusion-VPR) for large-scale place recognition problems \footnote{Supplementary Material: An accompanying video for this work is available at \url{https://youtu.be/g4tl--2nvhM}.}. FE-Fusion-VPR is an attention-based multi-scale deep network architecture for the mixed frame-/event-based VPR. It can effectively combine the advantages of standard frame cameras and event cameras, and suppress their disadvantages, thus achieving better VPR performance than using frames or events alone. To our knowledge, it is the first end-to-end deep network architecture combining standard cameras and event cameras for VPR problems and goes beyond the existing SOTA methods.
In summary, The main contributions of this paper are as follows:
\begin{itemize}
\item We propose a novel two-stream network (TSFE-Net) that can fuse intensity frame and event volume for hybrid feature extraction, which is compatible with asynchronous and irregular data from the event camera.
\item We propose an attention-based multi-scale network (MSF-Net) and design a re-weighting network (DRW-Net) that can assign weights to different sub-descriptors to obtain the best descriptor representation, achieving the SOTA VPR performance.
\item We comprehensively compare our FE-Fusion-VPR with other SOTA frame-/event-based VPR methods on the Brisbane-Event-VPR and DDD20 datasets to evaluate the advanced performance of our method.
\item We conduct ablation studies on each network component to comprehensively demonstrate the compactness of our FE-Fusion-VPR pipeline.
\end{itemize}

\section{Related Work}
\label{sec:relatedwork}

\subsection{Frame-based VPR Methods}

Conventional VPR algorithms mainly consist of two steps: feature extraction and feature matching. In feature extraction, key low-level features need to be extracted from high-dimensional visual data for storage. These key features are generally called descriptors or representations and can be local / global or sparse / dense.
Earlier frame-based VPR works focused on hand-crafted algorithms, including local feature extractors (SIFT \cite{lowe2004distinctive}, SURF \cite{bay2006surf}, ORB \cite{rublee2011orb}) and global feature extractors (HoG \cite{dalal2005histograms}, GiST \cite{oliva2006building}), feature clusterers (BoW \cite{angeli2008fast, galvez2012bags}, FV \cite{perronnin2010large, sanchez2013image}, VLAD \cite{jegou2010aggregating, arandjelovic2013all, torii201524, khaliq2019holistic}). However, hand-crafted methods usually need to be elaborately designed and are not robust to changes in illumination, seasons, viewpoint, etc. 
In contrast, learning-based (especially deep learning) feature extraction algorithms can learn general features and achieve better performance than hand-crafted algorithms in large-scale image retrieval tasks \cite{masone2021survey}. For example, Arandjelovic et al. improved VLAD as a trainable pooling layer (called NetVLAD \cite{arandjelovic2016netvlad}) for direct integration into the CNN-based VPR framework. 
On this basis, Patch-NetVLAD \cite{hausler2021patch} first computes the global NetVLAD descriptors to filter out some reference candidates, then uses the local patch NetVLAD descriptors for further fine-tuning matching, which obtains a higher performance than NetVLAD.
MR-NetVLAD \cite{khaliq2022multires} augments NetVLAD with multi-resolution image pyramid encoding, resulting in rich place representations to overcome challenging scenarios (such as illumination and viewpoint changes) better. 
Although learning-based VPR methods have achieved good performance, they still have difficulties in glare and high-speed scenes due to the inherent shortcomings of standard cameras.

\subsection{Event-based VPR Methods}
Recently, using event cameras to solve VPR problems has drawn more and more attention from researchers. First, Tobias et al. proposed an event-based VPR method (Ensemble-Event-VPR) with ensembles of temporal windows \cite{fischer2020event}, which reconstructs intensity frames from event streams of different temporal windows, extracts visual descriptors using NetVLAD \cite{arandjelovic2016netvlad}, and then integrates the distance matrix of multiple descriptors for VPR. Lee et al. proposed a VPR method (EventVLAD) to recover edge images from event streams and then use NetVLAD for descriptor generation \cite{lee2021eventvlad}. No matter whether using intensity frames or edge images, the above two methods need to transform event streams into frames. Therefore, they are still frame-based VPR methods in essence. To utilize event streams directly, we previously proposed an event-based end-to-end deep network architecture for VPR (Event-VPR) \cite{kong2022event}. Event-VPR uses EST voxel grid representation, combines deep residual network and VLAD layer to extract visual descriptors, and adopts weakly supervised loss for training, which achieves excellent performance in multiple challenging driving datasets using events directly. However, since event cameras can not output the appearance information of the scene directly, it is still tricky for VPR in weakly textured and motionless scenes.
The latest work, VEFNet \cite{huang2022vefnet}, simply uses a cross-modality attention module and a self-attention module for frame and events fusion based on the VGG network feature extractor. However, it does not perform better than NetVLAD-based VPR methods in most challenging cases. As a comparison, our proposed FE-Fusion-VPR outperforms both frame-based and event-based SOTA methods in performance.

\section{Methodology}
\label{sec:methodology}

\begin{figure*}[htbp]
\vspace{0mm}
\centering
\includegraphics[width=1\textwidth]{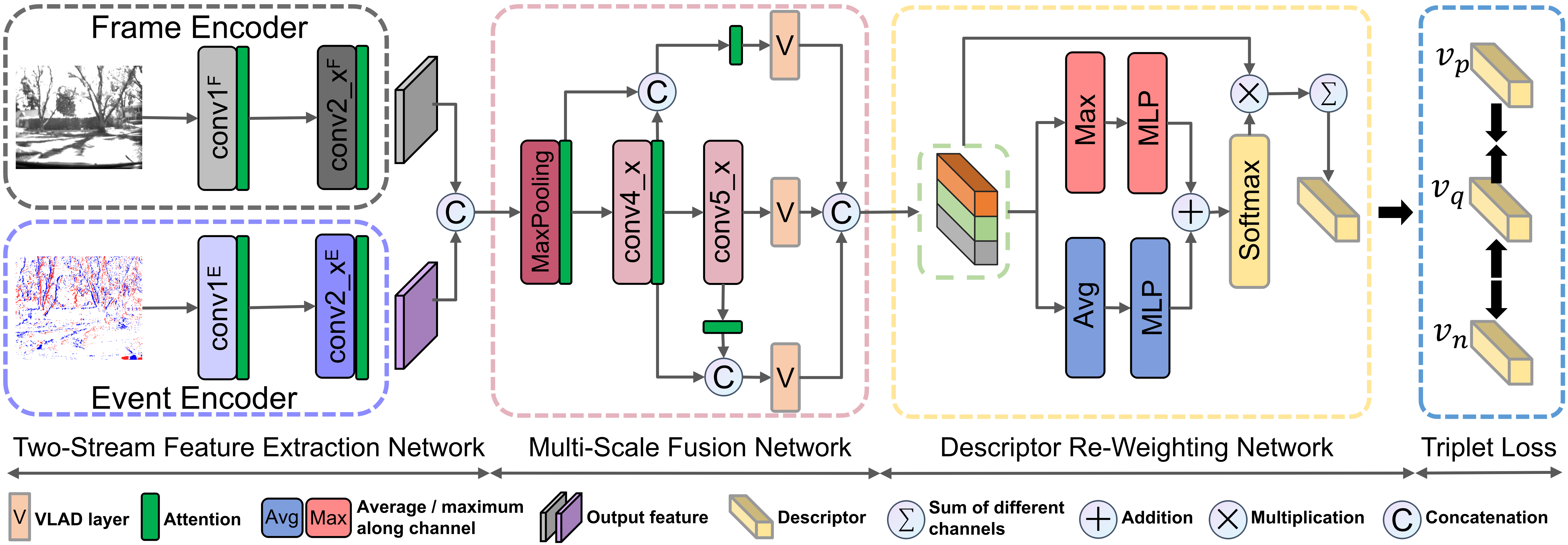}
\caption{Overview of the proposed FE-Fusion-VPR pipeline. First, the intensity frame and events are input into the two-stream feature extraction network for obtaining the fusion feature. Then, we leverage it as the input of the feature pyramid using the lateral connection for obtaining three-scale features and use the VLAD layer to get three corresponding sub-descriptoars for concatenation. Finally, we use descriptor re-weighting network to obtain the final refined descriptor for matching. The conv1, conv2\_x, conv4\_x and conv5\_x are convolutional layer and residual blocks cut from ResNet34.}
\label{fig:2}
\vspace{0mm}
\end{figure*}

\subsection{The Overall Architecture}
An overview of the proposed FE-Fusion-VPR pipeline is shown in Fig. \ref{fig:2}. Our FE-Fusion-VPR comprises a two-stream feature extraction network (TSFE-Net), a multi-scale fusion network (MSF-Net), and a descriptor re-weighting network (DRW-Net). Its backbone consists of residual blocks of ResNet34 \cite{he2016deep}. The intensity frame and events are first fed into TSFE-Net to obtain their shallow fusion feature. Then, MSF-Net extracts the multi-scale features which are aggregated to three sub-descriptors. Next, DRW-Net can learn the weights of the sub-descriptors for getting the final refined descriptor. Finally, the final descriptor is used to match the query data with reference data. The whole network is trained in an end-to-end manner with weakly supervision.
In addition, the attention layer is commonly used in our FE-Fusion-VPR network architecture, which consists of a channel attention mechanism and a spatial attention mechanism \cite{woo2018cbam}. We pass the feature maps of different scales through the attention layer to obtain more sophisticated feature maps, which can effectively improve VPR performance.

\subsection{Two-Stream Feature Extraction Network}

Due to the asynchronous characteristics of events, combining them with intensity frame remains challenging, especially for learning-based methods. In Event-VPR \cite{kong2022event}, our experiments have shown that different event representations have little effect on VPR tasks. Therefore in this paper, event volumes are processed to event frames \cite{maqueda2018event} directly, which are fed into our TSFE-Net together with intensity frames. As shown in Fig. \ref{fig:2}, in order to learn multi-modal shared feature, our two-stream feature extraction network (TSFE-Net) $\boldsymbol{f}_\text{TSFE-Net}(\cdot)$ extracts two kinds of shallow features and fuses them. Inspired by Event-VPR \cite{kong2022event}, we use convolutional layer conv1 and residual blocks conv2\_x cut from ResNet34, and attach an attention layer to each to improve the quality of shallow features.
For better performance, we concatenate the shallow features along the channel dimension \cite{ye2018hierarchical}. Finally, we use a max-pooling operation and an attention layer to obtain rich and effective scene information. Our two-stream feature extractor $\boldsymbol{f}_\text{TSFE}(\cdot)$ can be summarized as follows:
\begin{equation}
\label{eq:1}
\begin{aligned}
\boldsymbol{X}^\text{H}&=\boldsymbol{f}_\text{TSFE-Net}(\boldsymbol{F},\boldsymbol{E})   \\
&=\boldsymbol{f}_\mathrm{EC}^\text{F}(\boldsymbol{F}) \oplus \boldsymbol{f}_\mathrm{EC}^\text{E}(\boldsymbol{E})=\boldsymbol{X}^\text{F} \oplus \boldsymbol{X}^\text{E},
\end{aligned}
\end{equation}
where $\boldsymbol{f}_\text{EC}^\text{F}(\cdot)$ and $\boldsymbol{f}_\text{EC}^\text{F}(\cdot)$ are the encoders processing frames and events respectively. The encoder structure is Conv($7\mathrm{\times}7$,64,/2)-Attn-MaxPool2d(/2)-ResBlock0($3\mathrm{\times}3$,64)-ResBlock1($3\mathrm{\times}3$,64)-ResBlock2($3\mathrm{\times}3$,64)-Attn-BatchNorm-ReLU. $\boldsymbol{F}$ and $\boldsymbol{E}$ are intensity frame and event volume, $\boldsymbol{X}^\text{F}$ and $\boldsymbol{X}^\text{E}$ are the primary features of intensity frame and event volume respectively, $\oplus$ is the concatenation operation, and $\boldsymbol{X}^\text{H}$ is the hybrid feature after fusing.

\subsection{Multi-Scale Fusion Network}
\begin{figure}[htbp]
\vspace{0mm}
\centering
\includegraphics[width=\columnwidth]{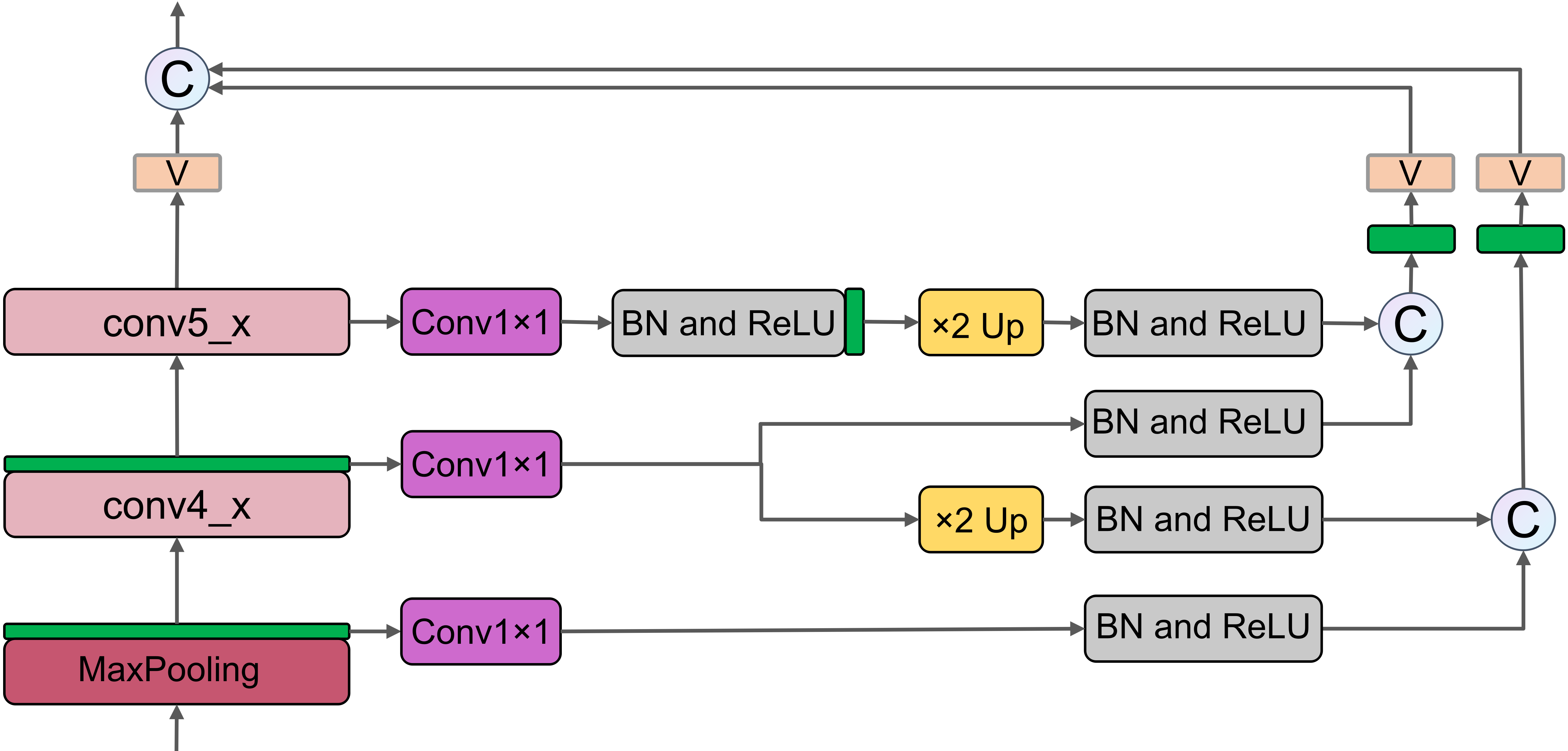}
\caption{Detailed illustration of our multi-scale fusion network (MSF-Net). In the MSF-Net, we obtain three-scale features and use the VLAD layer to get three corresponding sub-descriptors. Then, we concatenate them to obtain the primary multi-scale descriptor.}
\label{fig:3}
\vspace{0mm}
\end{figure}

Many works \cite{chen2014convolutional, sunderhauf2015performance} have demonstrated that mid-level visual features exhibit robustness to appearance changes, while high-level visual features are robust to viewpoint changes in VPR tasks. Therefore, the accuracy of VPR can be improved theoretically by using multi-scale network architecture. Here, our idea is inspired by feature pyramid network (FPN) \cite{lin2017feature}, which is a typical multi-scale network architecture. However, the performance of FPN degrades in some cases (such as the detection of large objects \cite{jin2022you}). To achieve more efficient communication between different levels, our proposed multi-scale fusion network (MSF-Net) $\boldsymbol{f}_\text{MSF-Net}(\cdot)$ fuses different-scale features in the following way (as shown in Fig. \ref{fig:3}). First, our backbone network performs bottom-up feature extraction, which contains three stages of residual structure. The output features of each stage are $\boldsymbol{S}_{1},\boldsymbol{S}_{2},\boldsymbol{S}_{3}$ respectively. In particular, the residual structure of the first stage is contained in $\boldsymbol{f}_\text{TSFE-Net}(\cdot)$. Thus, we directly adopt the max-pooling layer $\boldsymbol{f}_\text{MP}(\cdot)$ and the attention layer $\boldsymbol{f}_\text{Attn}(\cdot)$ for the output fusion feature $\boldsymbol{X}^\text{H}$ of the TSFE-Net to obtain the feature $\boldsymbol{S}_{1}$ of the first stage. Then, features $\boldsymbol{S}_{2}$ and $\boldsymbol{S}_{3}$ are extracted through the remaining two stages of residual structure (residual blocks conv4\_x and conv5\_x), which are expressed as follows:
\begin{equation}
\label{eq:2}
\begin{aligned}
\boldsymbol{S}_{1}&=\boldsymbol{f}_\text{Attn}(\boldsymbol{f}_\text{MP}(\boldsymbol{X}^\text{H})),   \\
\boldsymbol{S}_{2}&=\boldsymbol{f}_\text{Attn}(\boldsymbol{f}_\text{Res,2}(\boldsymbol{S}_{1})), \\
\boldsymbol{S}_{3}&=\boldsymbol{f}_\text{Res,3}(\boldsymbol{S}_{2}).
\end{aligned}
\end{equation}
where the channels of features $\boldsymbol{S}_{1},\boldsymbol{S}_{2},\boldsymbol{S}_{3}$ are $128,256,512$ respectively.
Next, different from FPN, to make more efficient use of the multi-scale information of each stage, we adopt concatenation to perform stage-wise fusion based on the backbone network.
Specifically, for each stage of the backbone network, we add the branch network as a lateral connection (passway) to fuse the features of the two adjacent stages. The branch network generally includes a convolutional layer (convolution kernel is 1×1), upsampling layer (x2 Up), batch-normalization layer (BN), and ReLU activation layer.
By adjusting the channels and spatial resolution of features, we can obtain features $\boldsymbol{M}_{1},\boldsymbol{M}_{2},\boldsymbol{M}_{3}$ with 256 channels and $8\times8$, $16\times16$ and $32\times32$ spatial resolutions respectively:
\begin{equation}
\label{eq:3}
\boldsymbol{M}_{1},\boldsymbol{M}_{2},\boldsymbol{M}_{3}=\boldsymbol{f}_\text{MSF-Net}(\boldsymbol{X}^\text{H}).
\end{equation}
After that, we use the VLAD layer \cite{arandjelovic2016netvlad, kong2022event} $\boldsymbol{f}_\text{VLAD}(\cdot)$ for three features respectively to get three corresponding sub-descriptors $\{\boldsymbol{D}_1,\boldsymbol{D}_2,\boldsymbol{D}_3\}$, whose dimensions are all $1\times\mathrm{N}$. Next, we concatenate them to obtain the primary multi-scale descriptor $\widetilde{\boldsymbol{D}}$ of which dimension is $3\times\mathrm{N}$:
\begin{equation}
\label{eq:4}
\begin{aligned}
\boldsymbol{D}&=\mathop{\Big{\|}}_{i=1}^3\left(\boldsymbol{D}_{i}\right)&=\mathop{\Big{\|}}_{i=1}^3\left(\boldsymbol{f}_\text{VLAD}\left(\boldsymbol{M}_{i}\right)\right),
\end{aligned}
\end{equation}
where $\|(\cdot)$ represents concatenation operation. 
Since the multi-scale fusion features contain rich scenario details and powerful semantic features, they can provide robust feature information for the DRW-Net to improve our network's overall performance.

\subsection{Descriptor Re-Weighting Network}

\begin{figure}[htbp]
\vspace{-0mm}
\centering
\includegraphics[width=\columnwidth]{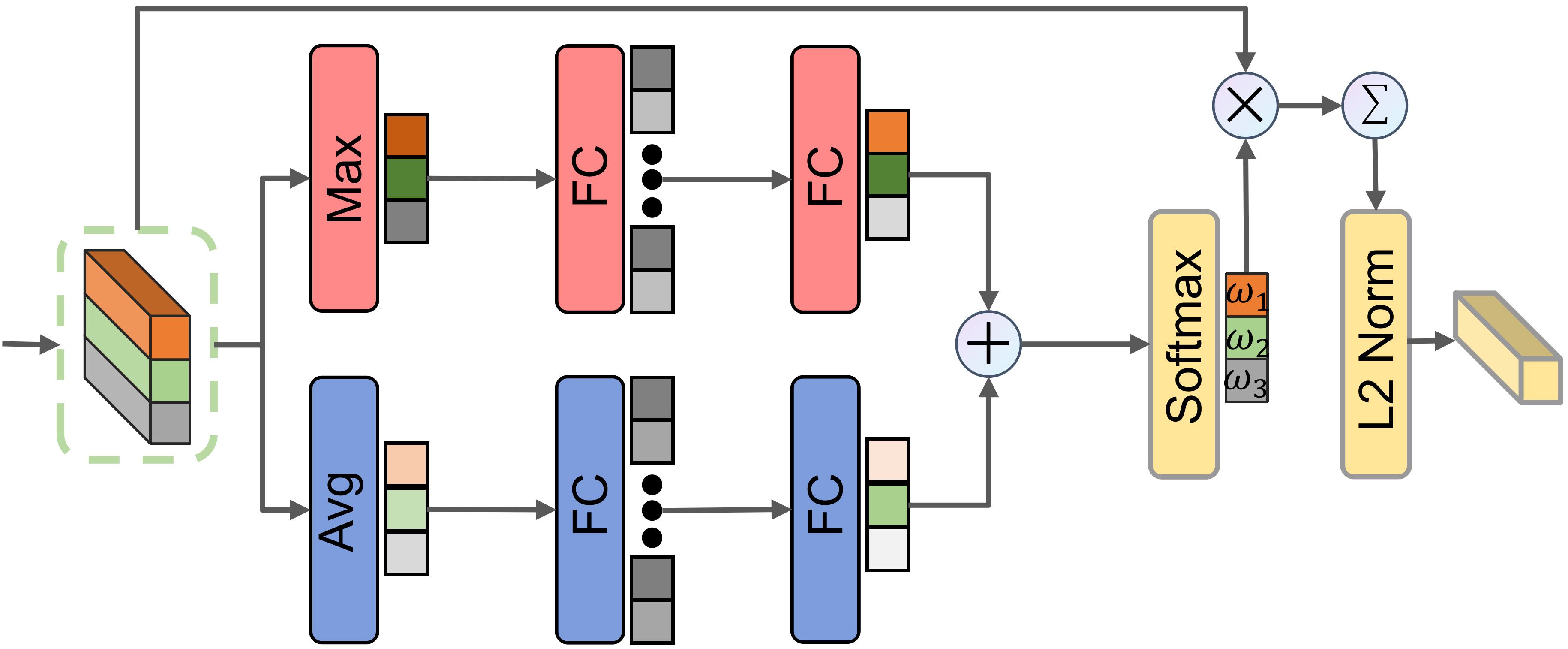}
\caption{Detailed illustration of our descriptor re-weighting network (DRW-Net). We use maximum, average operation, MLP network and soft-max layer to obtain the weights of the primary sub-descriptors. And then, we utilize the weights to get the final refined descriptor.}
\label{fig:4}
\vspace{0mm}
\end{figure}

In MSF-Net, we have obtained the primary multi-scale descriptor aggregated by three different-scale features. In order to represent the scenes better, we need to redesign the multi-scale descriptor to obtain a compact global descriptor. Therefore, we propose a descriptor re-weighting network (DRW-Net) $\boldsymbol{f}_\text{DRW-Net}(\cdot)$, as shown in Fig. \ref{fig:4}, to obtain a robust global descriptor:
\begin{equation}
\label{eq:5}
\boldsymbol{D}^{\prime}=\boldsymbol{f}_\text{DRW-Net}\left(\boldsymbol{D}\right).
\end{equation}
For the multi-scale descriptor $\boldsymbol{D}$, we calculate the average $\boldsymbol{f}_\text{avg}(\cdot)$ and maximum $\boldsymbol{f}_\text{max}(\cdot)$ of each sub-descriptor respectively:
\begin{equation}
\label{eq:6}
\begin{aligned}
\boldsymbol{G}_\text{avg}&=\boldsymbol{f}_\text{avg}(\boldsymbol{D})=\mathop{\Big{\|}}_{i=1}^3\left(\frac{1}{\mathrm{N}} \sum_{m=1}^\mathrm{N} {\boldsymbol{D}_i(m)}\right),  \\
\boldsymbol{G}_\text{max}&=\boldsymbol{f}_\text{max}(\boldsymbol{D})=\mathop{\Big{\|}}_{i=1}^3\left(\max_{m} \left(\boldsymbol{D}_i(m)\right)\right),
\end{aligned}
\end{equation}
where $\boldsymbol{G}_\text{avg}$ and $\boldsymbol{G}_\text{max}$ are the channel-wise global representations of sub-descriptors, $\mathrm{N}$ denotes the index of the sub-descriptor, $m$ is the spatial coordinates of the sub-descriptors. Then, we append two fully connected (FC) layers respectively to learn two kinds of weights of the sub-descriptors $\boldsymbol{w}_\text{avg}$ and $\boldsymbol{w}_\text{max}$, and add the above two weights to obtain the final weights $\boldsymbol{w}$ of sub-descriptors through a soft-max layer $\boldsymbol{f}_\text{SM}(\cdot)$:
\begin{equation}
\label{eq:7}
\begin{aligned}
\boldsymbol{w}_\text{avg}&=\boldsymbol{f}_{\text{FC},2}\left(\boldsymbol{f}_\text{ReLU}\left(\boldsymbol{f}_{\text{FC},1}\left(\boldsymbol{G}_\text{max}\right)\right)\right),  \\
\boldsymbol{w}_\text{max}&=\boldsymbol{f}_{\text{FC},2}^{\prime}\left(\boldsymbol{f}_\text{ReLU}\left(\boldsymbol{f}_{\text{FC},1}^{\prime}\left(\boldsymbol{G}_\text{max}\right)\right)\right),  \\
\boldsymbol{w}&=\boldsymbol{f}_\text{SM}\left(\boldsymbol{w}_\text{avg}+\boldsymbol{w}_\text{max}\right),
\end{aligned}
\end{equation}
where $\boldsymbol{f}_{\text{FC},1}(\cdot),\boldsymbol{f}_{\text{FC},1}^{\prime}(\cdot):\mathbb{R}^{3\times1}\rightarrow\mathbb{R}^{\mathrm{M}\times1}$ and $\boldsymbol{f}_{\text{FC},2}(\cdot),\boldsymbol{f}_{\text{FC},2}^{\prime}(\cdot):\mathbb{R}^{\mathrm{M}\times1}\rightarrow\mathbb{R}^{3\times1}$ are two kinds of fully connected (FC) layers respectively. $\mathrm{M}=12$ is the transformation length of global representations in hidden layers.
The soft-max operation $\boldsymbol{f}_\text{SM}(\cdot)$ makes the weights of these three descriptors mutually balanced, and the sum of their weights is 1. Finally, we multiply the multi-scale descriptors with weights and sum multi-scale descriptors channel by channel, which is expressed as follows:
\begin{equation}
\label{eq:8}
\begin{aligned}
\boldsymbol{D}^{\prime} &=\boldsymbol{D} \otimes \boldsymbol{w} \\
&=\boldsymbol{D}_1 \otimes w_{1} + \boldsymbol{D}_2 \otimes w_{2} + \boldsymbol{D}_3 \times w_{3}   \\
&=\boldsymbol{D}_1^{\prime} + \boldsymbol{D}_2^{\prime} + \boldsymbol{D}_3^{\prime} .
\end{aligned}
\end{equation}
Thus, we obtain the final multi-scale weighted aggregate descriptor $\boldsymbol{D}^{\prime}$. It is a more robust place representation than the primary descriptor $\boldsymbol{D}$.

\begin{table*}[htbp]
\vspace{-0mm}
\begin{center}
\caption{Scenarios, sequences, training and testing sets of Brisbane-Event-VPR \cite{fischer2020event} and DDD20 \cite{hu2020ddd20} datasets used in our experiments.}
\newcommand{\tabincell}[2]{\begin{tabular}{@{}#1@{}}#2\end{tabular}}
\small
\setlength{\tabcolsep}{2mm}{
\begin{tabular}{|c|c|c|c|p{1cm}}
\hline
\multirow{2}{*}{Datasets}
&\multirow{2}{*}{Scenarios \& Sequences}
&\multirow{2}{*}{\tabincell{c}{Training Sets\\(Database / Query)}}
&\multirow{2}{*}{\tabincell{c}{Testing Sets\\(Database / Query)}} \\
&   &   &   \\
\hline
\multirow{4}{*}{\tabincell{c}{Brisbane- \\ Event-VPR \\ \cite{fischer2020event}}}
&\multirow{4}{*}{\tabincell{c}{sunrise (sr) [2020-04-29-06-20-23] \\ morning (mn) [2020-04-28-09-14-11] \\ datime (dt) [2020-04-24-15-12-03] \\ sunset (ss) [2020-04-21-17-03-03][2020-04-22-17-24-21]}}
&(dt \& mn) [4712] / sr [2620]  &ss1 [1768] / ss2 [1768]    \\
\cline{3-4}
&   &(ss2 \& mn) [4246] / sr [2620] &ss1 [2492] / dt [2492] \\
\cline{3-4}
&   &(ss2 \& dt) [4002] / sr [2620] &ss1 [2478] / mn [2478] \\
\cline{3-4}
&   &(ss2 \& dt) [4002] / mn [2478] &ss1 [2181] / sr [2181] \\
\hline
\multirow{2}{*}{\tabincell{c}{DDD20 \\ \cite{hu2020ddd20}}}
&\multirow{2}{*}{\tabincell{c}{street [rec1501983083][rec1502648048][rec1502325857] \\ freeway [rec1500924281][rec1501191354][rec1501268968]}} 
&**83 [4246] / **57 [2620]  &**83 [2492] / **48 [2492]  \\
\cline{3-4}
&   &**81 [4712] / **54 [2620]  &**81 [1768] / **68 [1768]  \\
\hline
\end{tabular}}
\label{tab:1}
\vspace{-5mm}
\end{center}
\end{table*}

\section{Experiments}
\label{sec:experiments}
\subsection{Experimental Setup}
\subsubsection{Dataset Selection}
To evaluate the performance of the proposed method, we conduct experiments on Brisbane-Event-VPR \cite{fischer2020event} and DDD20 \cite{hu2020ddd20} datasets. Brisbane-Event-VPR \cite{fischer2020event} consists of data recorded using a DAVIS camera together with GPS. It includes six traverses of the same path at different time of the day, including sunrise, morning, daytime, sunset, and night. We discard the night sequence due to too sparse intensity frames. DDD20 \cite{hu2020ddd20} is the event camera end-to-end driving dataset under various lighting conditions. We selected six sequences of two urban scenes, of which two sequences have glare illumination, and three sequences consist of highways. We use the timestamps of the intensity frames to get the event volumes. The intervals of Brisbane-Event-VPR and DDD20 datasets that we select are approximately 25ms and 20ms respectively.

\begin{figure}[htbp]
\centering
\subfigure[]{\label{fig:5a}\includegraphics[width=\columnwidth]{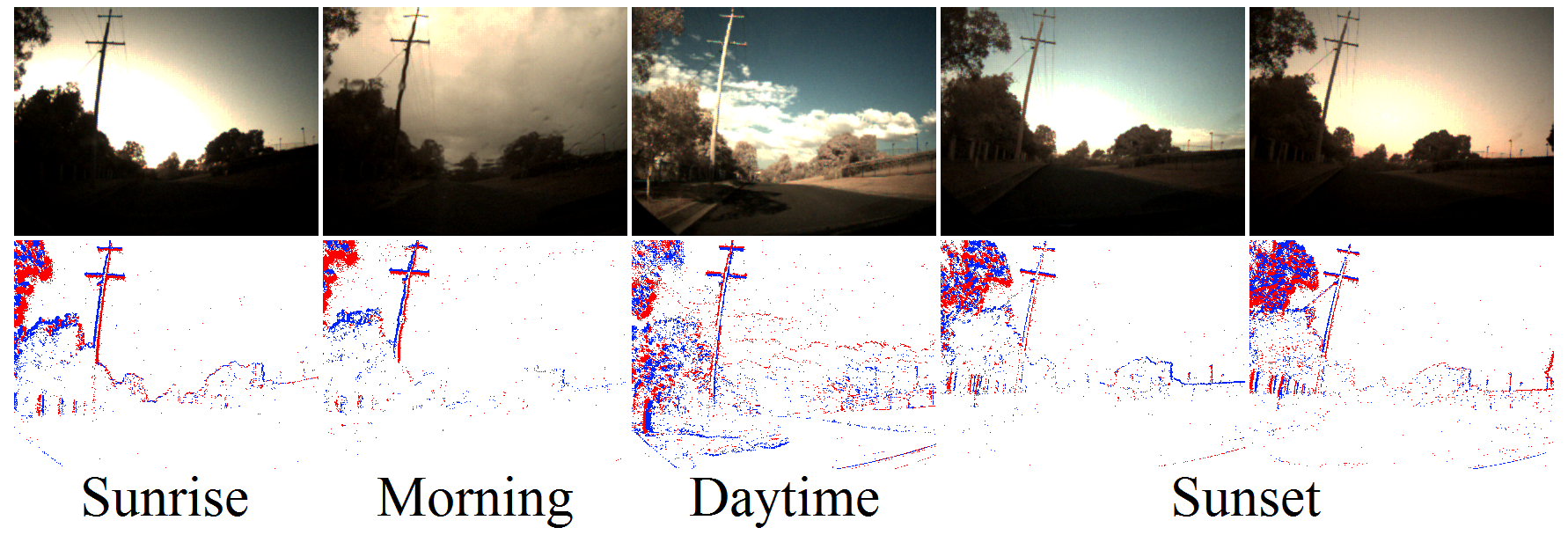}}
\subfigure[]{\label{fig:5b}\includegraphics[width=\columnwidth]{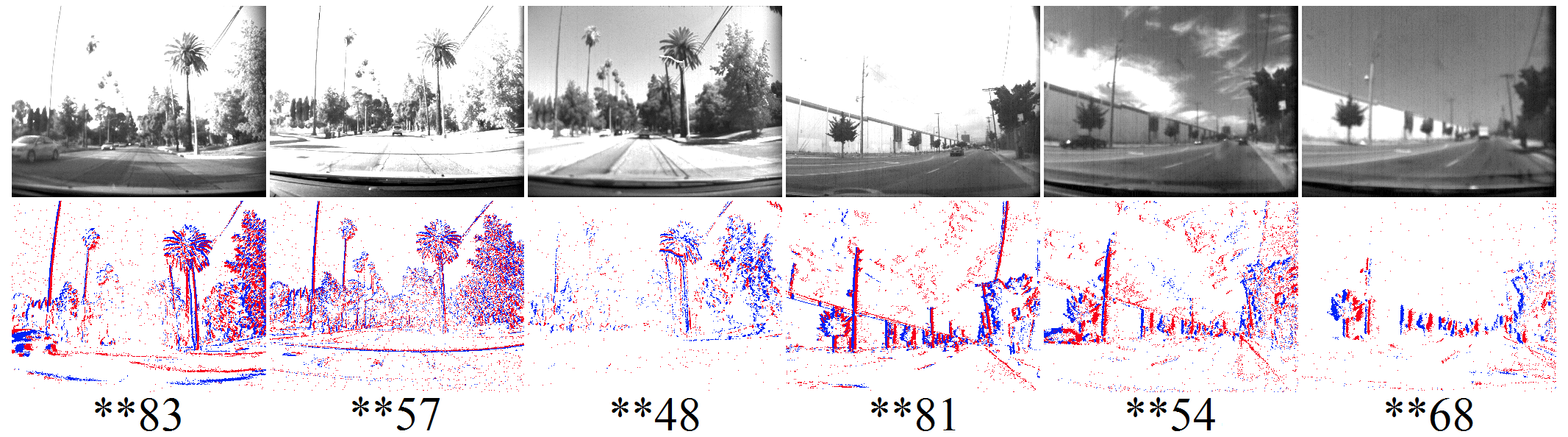}}
\caption{Scenarios of Brisbane-Event-VPR \cite{fischer2020event} and DDD20 \cite{hu2020ddd20} datasets in our experiments. (a) Brisbane-Event-VPR \cite{fischer2020event}. (b) DDD20 \cite{hu2020ddd20}.}
\label{fig:5}
\vspace{-0mm}
\end{figure}

\subsubsection{Parameters in Training}
We train our FE-Fusion-VPR network with weakly supervision using a triplet ranking loss. Except for the optimizer and learning rate, we use the same parameters in all experiments for a fair comparison. Where, the number of cluster centers (vocabulary size) $\mathrm{K}=128$, margin $\varepsilon=0.1$. We choose the potential positive distance threshold $\lambda=25\text{m}$, the randomly negative distance threshold $\delta=75\text{m}$ and the true positive geographical distance threshold $\phi=75\text{m}$. For more details of the training process, please directly refer to Event-VPR \cite{kong2022event}.

\subsubsection{Evaluation Metrics}
We use PR curve and Recall@N to evaluate the experimental results, which you can refer to Event-VPR \cite{kong2022event} for the specific description of the metrics. For a more comprehensive comparison, we calculate F1-max for all the VPR methods, which is as follows:
\begin{equation}
\label{eq:9}
F_{1,\text{max}} = \max_k{\left(F_{1,k}\right)} = \max_k{\left(\frac{2 \times P_k \times R_k}{P_k + R_k}\right)},
\end{equation}
where $P_k$ and $R_k$ are the $k$-th precision and recall in PR curves respectively. In addition, we also present the retrieval success-rate maps to show the performance of our algorithm more intuitively.

\begin{table*}[htbp]
\vspace{0mm}
\begin{center}
\caption{Comparison of our FE-Fusion-VPR against frame-based SOTA VPR methods \cite{arandjelovic2016netvlad,khaliq2022multires} and event-based SOTA VPR methods \cite{fischer2020event} \cite{kong2022event} on Brisbane-Event-VPR \cite{fischer2020event} and DDD20 \cite{hu2020ddd20} datasets with the best result \textbf{bolded}.}
\setlength{\tabcolsep}{1pt}
\newcommand{\tabincell}[2]{\begin{tabular}{@{}#1@{}}#2\end{tabular}}
\small
\setlength{\tabcolsep}{0.01\linewidth}
\begin{tabular}{|c|c|c|c|c|c|c|c|p{1cm}}
\cline{1-8}
\multirow{4}{*}{\tabincell{c}{VPR \\ Methods}}
&\multirow{4}{*}{\tabincell{c}{Evaluation \\ Metrics}}
&\multicolumn{6}{c|}{\tabincell{c}{Training Set \& Testing Set (database / query)}}  \\
\cline{3-8}
&   &\multicolumn{4}{c|}{\tabincell{c}{Brisbane-Event-VPR \cite{fischer2020event} Dataset}}
&\multicolumn{2}{c|}{\tabincell{c}{DDD20 \cite{hu2020ddd20} Dataset}}  \\
\cline{3-8}
&   &\multirow{2}{*}{\tabincell{c}{(dt \& mr) / sr \\ ss1 / ss2}} 
&\multirow{2}{*}{\tabincell{c}{(ss2 \& dt) / mr \\ ss1 / sr}} 
&\multirow{2}{*}{\tabincell{c}{(ss2 \& dt) / sr \\ ss1 / mr}} 
&\multirow{2}{*}{\tabincell{c}{(ss2 \& mr) / sr \\ ss1 / dt}} 
&\multirow{2}{*}{\tabincell{c}{**83 / **57 \\ **83 / **48}}
&\multirow{2}{*}{\tabincell{c}{**81 / **54 \\ **81 / **68}} \\
&   &   &   &   &   &   &   \\
\cline{1-8}
\multirow{2}{*}{\tabincell{c}{NetVLAD \\ \cite{arandjelovic2016netvlad}}}
&\multirow{2}{*}{\tabincell{c}{Recall@1/5 (\%) \\ F1-max $\uparrow$}}
&\multirow{2}{*}{\tabincell{c}{94.34, 97.29 \\ 0.9709}}
&\multirow{2}{*}{\tabincell{c}{90.61, 96.35 \\ 0.9546}}
&\multirow{2}{*}{\tabincell{c}{86.97, 94.11 \\ 0.9330}}
&\multirow{2}{*}{\tabincell{c}{77.40, 88.49 \\ 0.8762}}
&\multirow{2}{*}{\tabincell{c}{46.08, 62.91 \\ 0.6383}}
&\multirow{2}{*}{\tabincell{c}{9.30, 17.89 \\ 0.2411}}   \\
&   &   &   &   &   &   &   \\
\cline{1-8}
\multirow{2}{*}{\tabincell{c}{MR-NetVLAD \\ \cite{khaliq2022multires}}}
&\multirow{2}{*}{\tabincell{c}{Recall@1/5  (\%) \\ F1-max $\uparrow$}}
&\multirow{2}{*}{\tabincell{c}{94.23, 97.06 \\ 0.9733}}
&\multirow{2}{*}{\tabincell{c}{91.46, 95.43 \\ 0.9582}}
&\multirow{2}{*}{\tabincell{c}{87.20, 93.06 \\ \textbf{0.9343}}}
&\multirow{2}{*}{\tabincell{c}{79.68, 88.72 \\ 0.8929}}
&\multirow{2}{*}{\tabincell{c}{64.66, 82.55 \\ 0.7860}}
&\multirow{2}{*}{\tabincell{c}{30.38, 43.56 \\ 0.5244}}   \\
&   &   &   &   &   &   &   \\
\cline{1-8}
\multirow{2}{*}{\tabincell{c}{Ensemble-Event-\\VPR \cite{fischer2020event}}}
&\multirow{2}{*}{\tabincell{c}{Recall@1/5 (\%) \\ F1-max $\uparrow$}}
&\multirow{2}{*}{\tabincell{c}{87.33, 95.70 \\ 0.9345}}
&\multirow{2}{*}{\tabincell{c}{58.82, 82.42 \\ 0.7246}}
&\multirow{2}{*}{\tabincell{c}{58.42, 78.33 \\ 0.7550}}
&\multirow{2}{*}{\tabincell{c}{43.62, 62.71 \\ 0.6319}}
&\multirow{2}{*}{\tabincell{c}{32.02, 50.17 \\ 0.4967}}
&\multirow{2}{*}{\tabincell{c}{7.84, 14.99 \\ 0.1465}}   \\
&   &   &   &   &   &   &   \\
\cline{1-8}
\multirow{2}{*}{\tabincell{c}{Event-VPR \\ (Ours) \cite{kong2022event}}}
&\multirow{2}{*}{\tabincell{c}{Recall@1/5 (\%) \\ F1-max $\uparrow$}}
&\multirow{2}{*}{\tabincell{c}{84.79, 93.83 \\ 0.9236}}
&\multirow{2}{*}{\tabincell{c}{65.65, 86.52 \\ 0.7984}}
&\multirow{2}{*}{\tabincell{c}{66.67, 84.26 \\ 0.7946}}
&\multirow{2}{*}{\tabincell{c}{44.54, 66.10 \\ 0.6705}}
&\multirow{2}{*}{\tabincell{c}{43.89, 66.67 \\ 0.6057}}
&\multirow{2}{*}{\tabincell{c}{8.52, 21.33 \\ 0.1578}}   \\
&   &   &   &   &   &   &   \\
\cline{1-8}
\multirow{2}{*}{\tabincell{c}{\textbf{FE-Fusion-}\\\textbf{VPR (Ours)}}}
&\multirow{2}{*}{\tabincell{c}{Recall@1/5 (\%) \\ F1-max $\uparrow$}}
&\multirow{2}{*}{\tabincell{c}{\textbf{95.64, 98.36} \\ \textbf{0.9780}}}
&\multirow{2}{*}{\tabincell{c}{\textbf{93.58, 97.19} \\ \textbf{0.9671}}}
&\multirow{2}{*}{\tabincell{c}{\textbf{87.41, 93.87} \\ 0.9310}}
&\multirow{2}{*}{\tabincell{c}{\textbf{86.15, 93.03} \\ \textbf{0.9247}}}
&\multirow{2}{*}{\tabincell{c}{\textbf{72.77, 86.04} \\ \textbf{0.8407}}}
&\multirow{2}{*}{\tabincell{c}{\textbf{54.05, 64.34} \\ \textbf{0.7465}}}   \\
&   &   &   &   &   &   &   \\
\cline{1-8}
\end{tabular}
\label{tab:3}
\vspace{0mm}
\end{center}
\end{table*}

\begin{figure*}[htbp]
\vspace{-0mm}
\centering
\subfigure[sunset1 / sunset2]{\label{fig:6a}\includegraphics[width=0.16\textwidth]{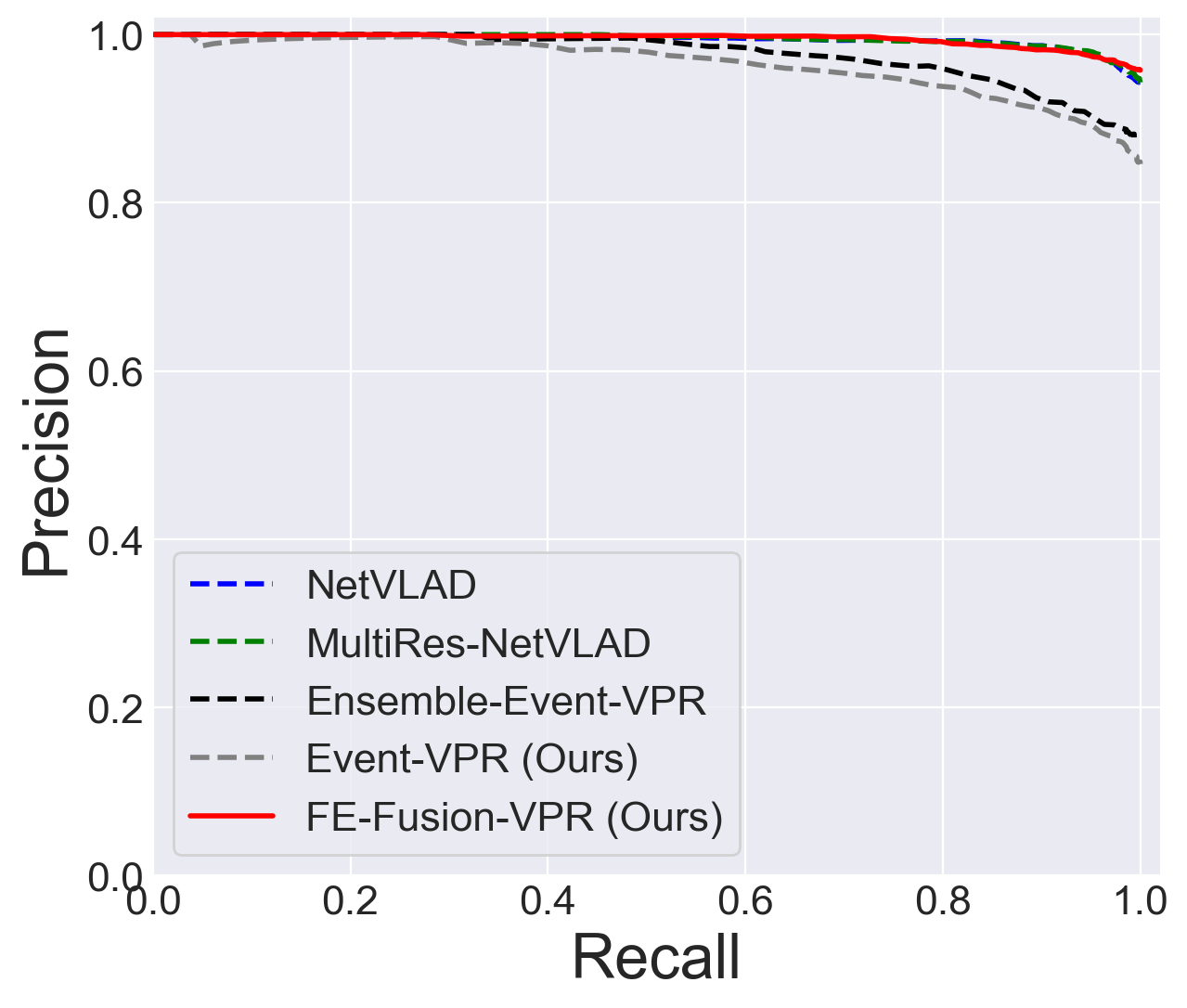}}
\subfigure[sunset1 / sunrise]{\label{fig:6b}\includegraphics[width=0.16\textwidth]{fig_6_a.png}}
\subfigure[sunset1 / morning]{\label{fig:6c}\includegraphics[width=0.16\textwidth]{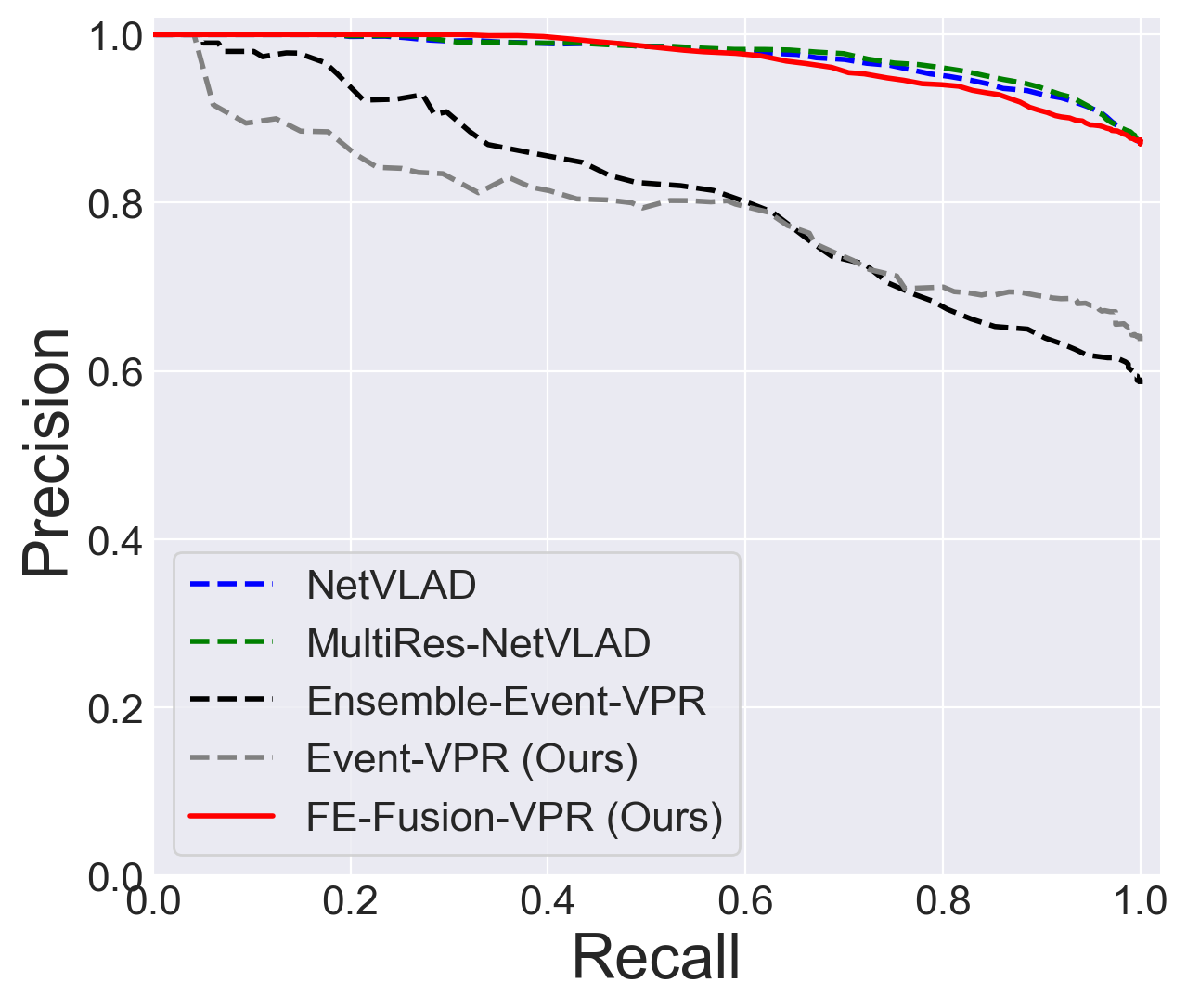}}
\subfigure[sunset1 / daytime]{\label{fig:6d}\includegraphics[width=0.16\textwidth]{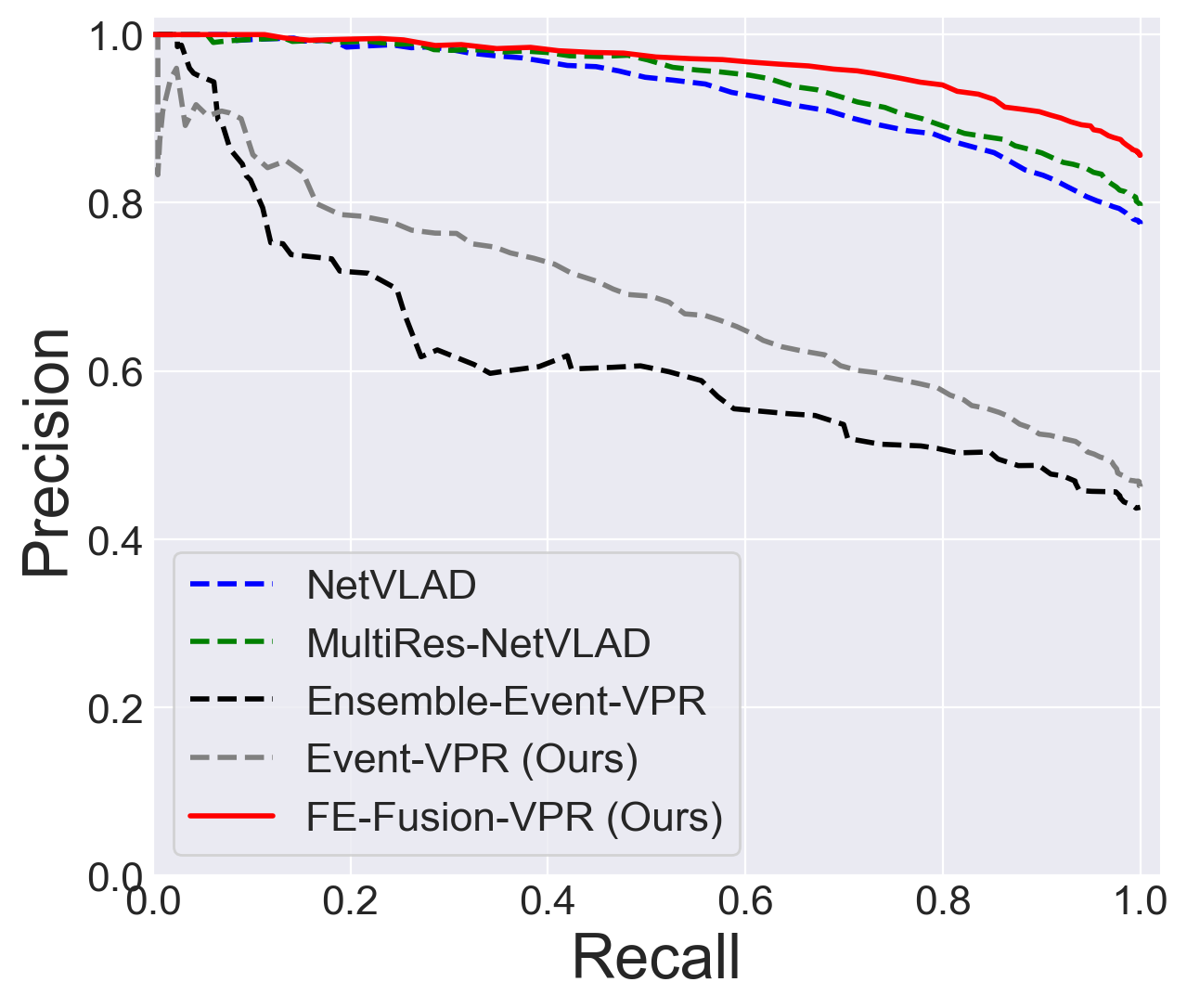}}
\subfigure[**83 / **48]{\label{fig:6e}\includegraphics[width=0.16\textwidth]{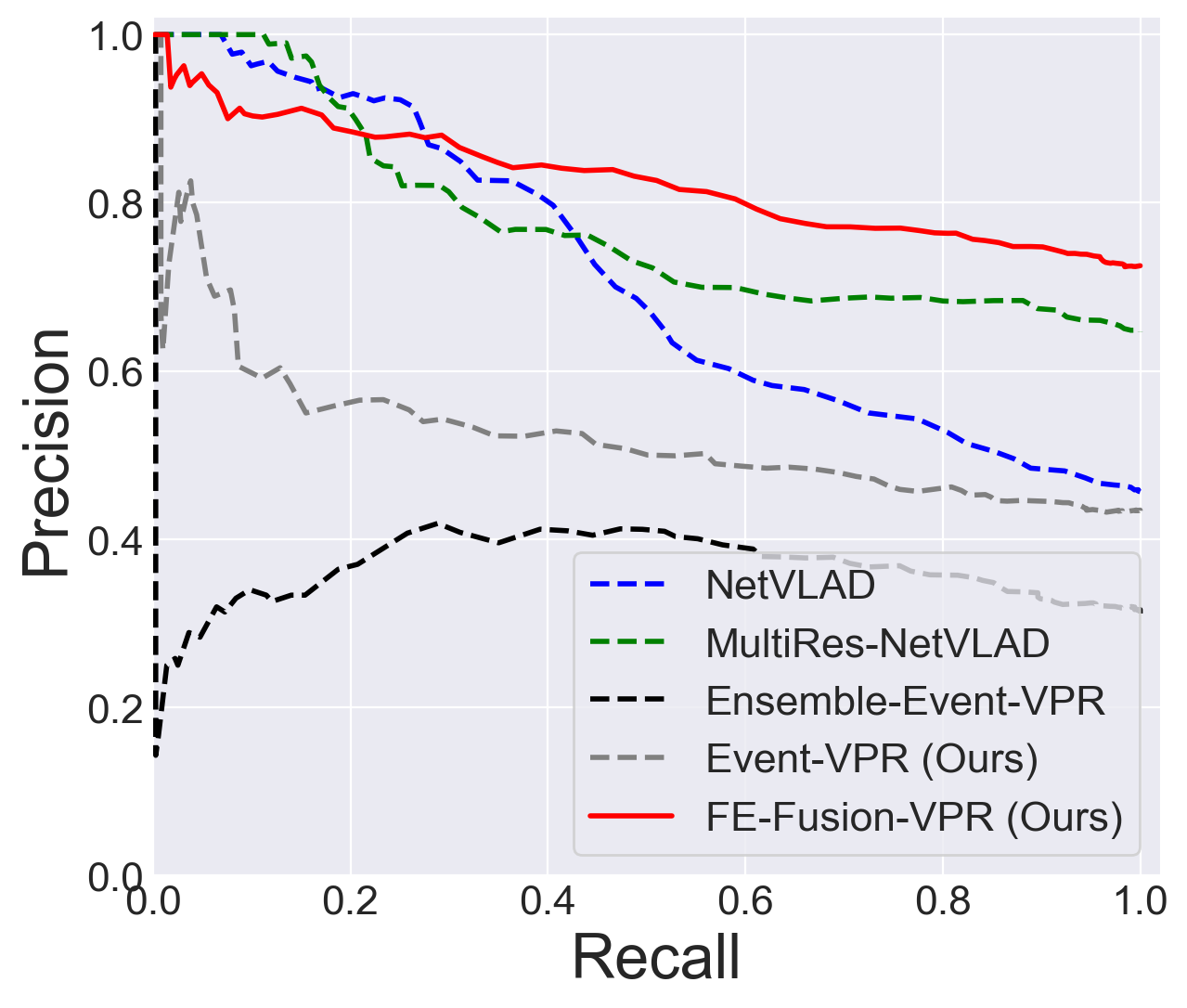}}
\subfigure[**81 / **68]{\label{fig:6f}\includegraphics[width=0.16\textwidth]{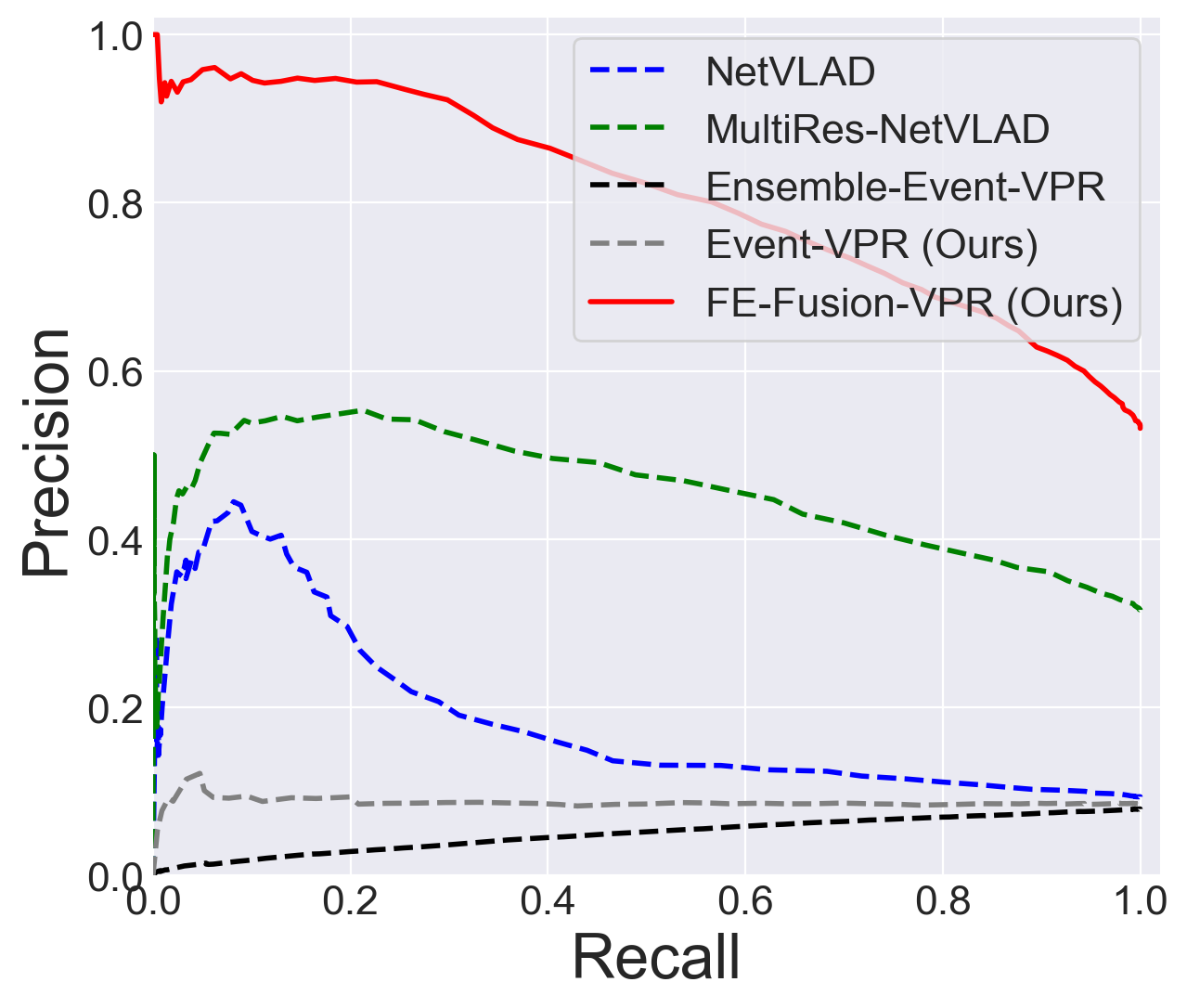}}
\caption{PR curves of NetVLAD \cite{arandjelovic2016netvlad}, MR-NetVLAD \cite{khaliq2022multires}, Ensemble-Event-VPR \cite{fischer2020event}, Event-VPR (ours) \cite{kong2022event} and FE-Fusion-VPR (ours) on Brisbane-Event-VPR \cite{fischer2020event} and DDD20 \cite{hu2020ddd20} datasets. (a)-(d) are on Brisbane-Event-VPR \cite{fischer2020event} dataset, and (e), (f) are on DDD20 \cite{hu2020ddd20} dataset. Our FE-Fusion-VPR (red) performs better than SOTA methods under most scenes.}
\label{fig:6}
\vspace{-2mm}
\end{figure*}

\begin{figure*}[htbp]
\vspace{-0mm}
\centering
\subfigure[sunset1 / sunset2]{\label{fig:7a}\includegraphics[width=0.16\textwidth]{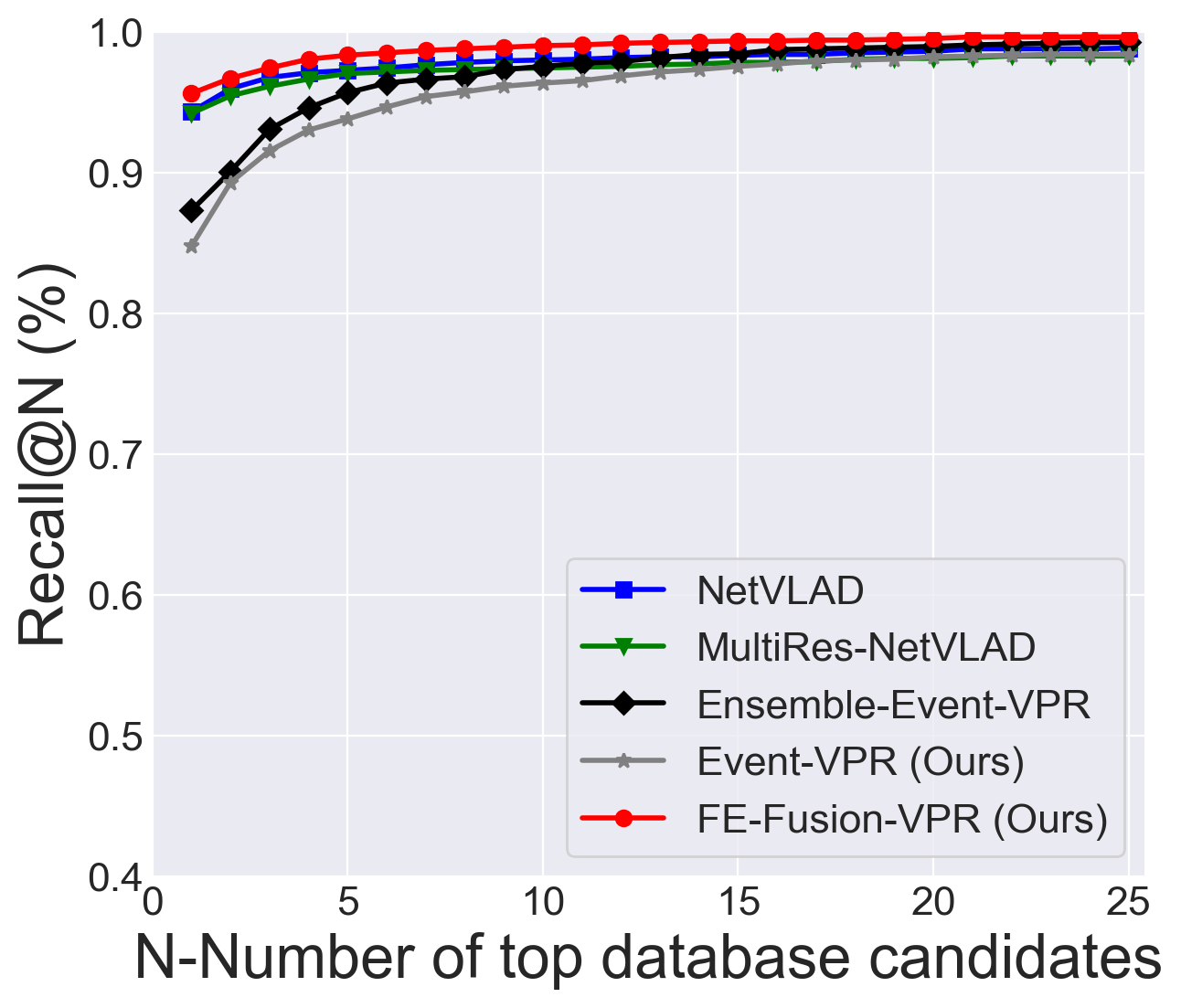}}
\subfigure[sunset1 / sunrise]{\label{fig:7b}\includegraphics[width=0.16\textwidth]{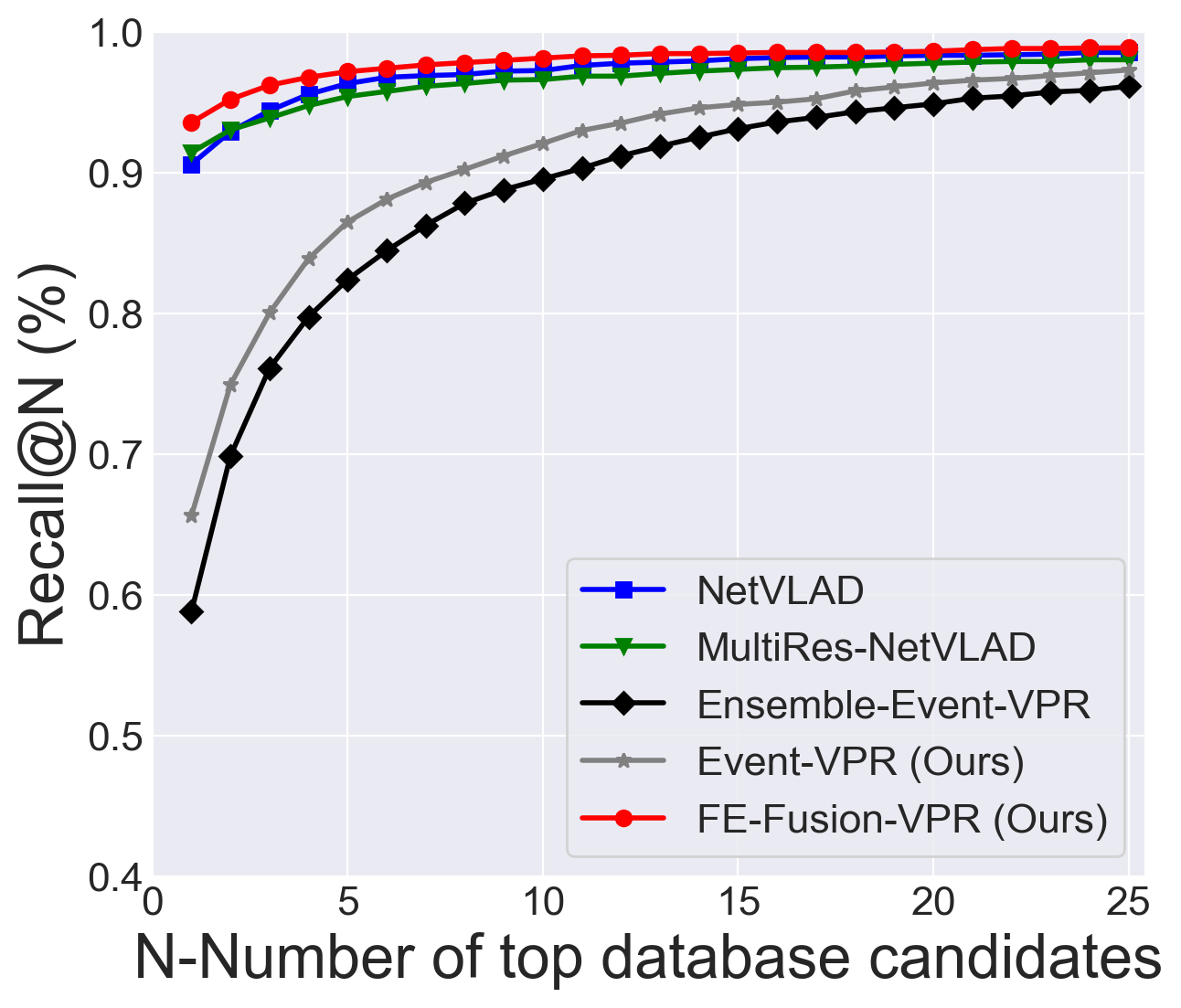}}
\subfigure[sunset1 / morning]{\label{fig:7c}\includegraphics[width=0.16\textwidth]{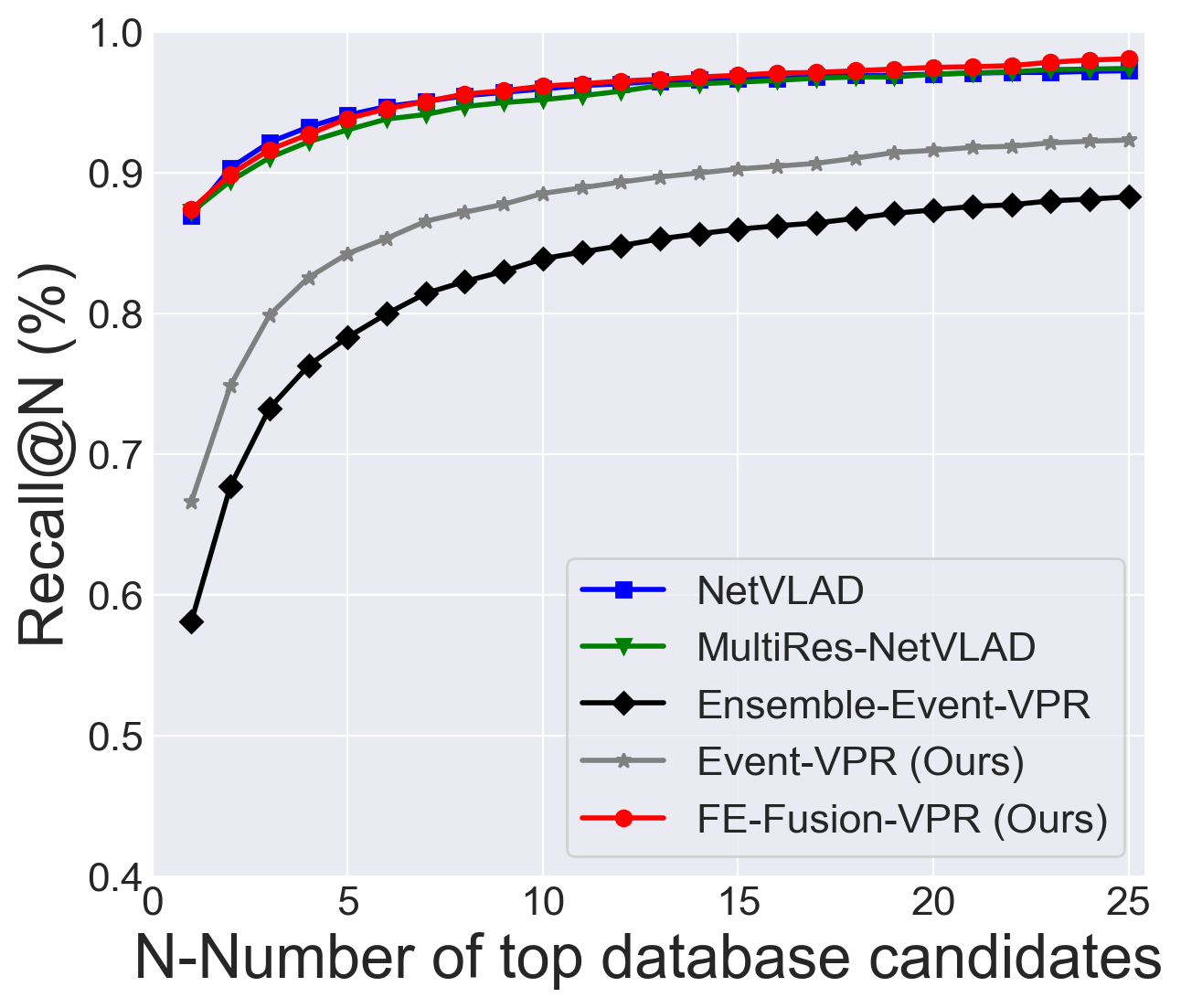}}
\subfigure[sunset1 / daytime]{\label{fig:7d}\includegraphics[width=0.16\textwidth]{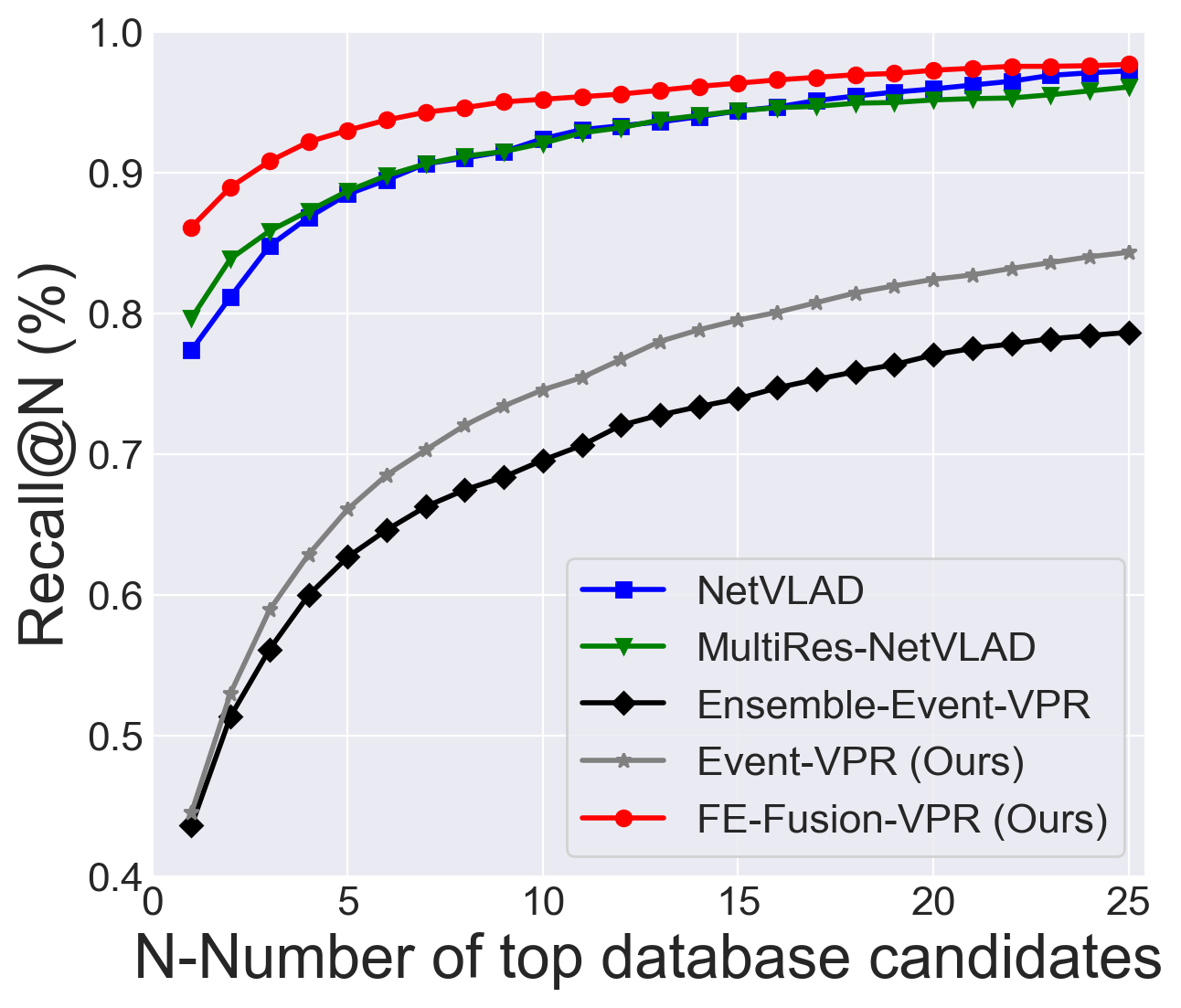}}
\subfigure[**83 / **48]{\label{fig:7e}\includegraphics[width=0.16\textwidth]{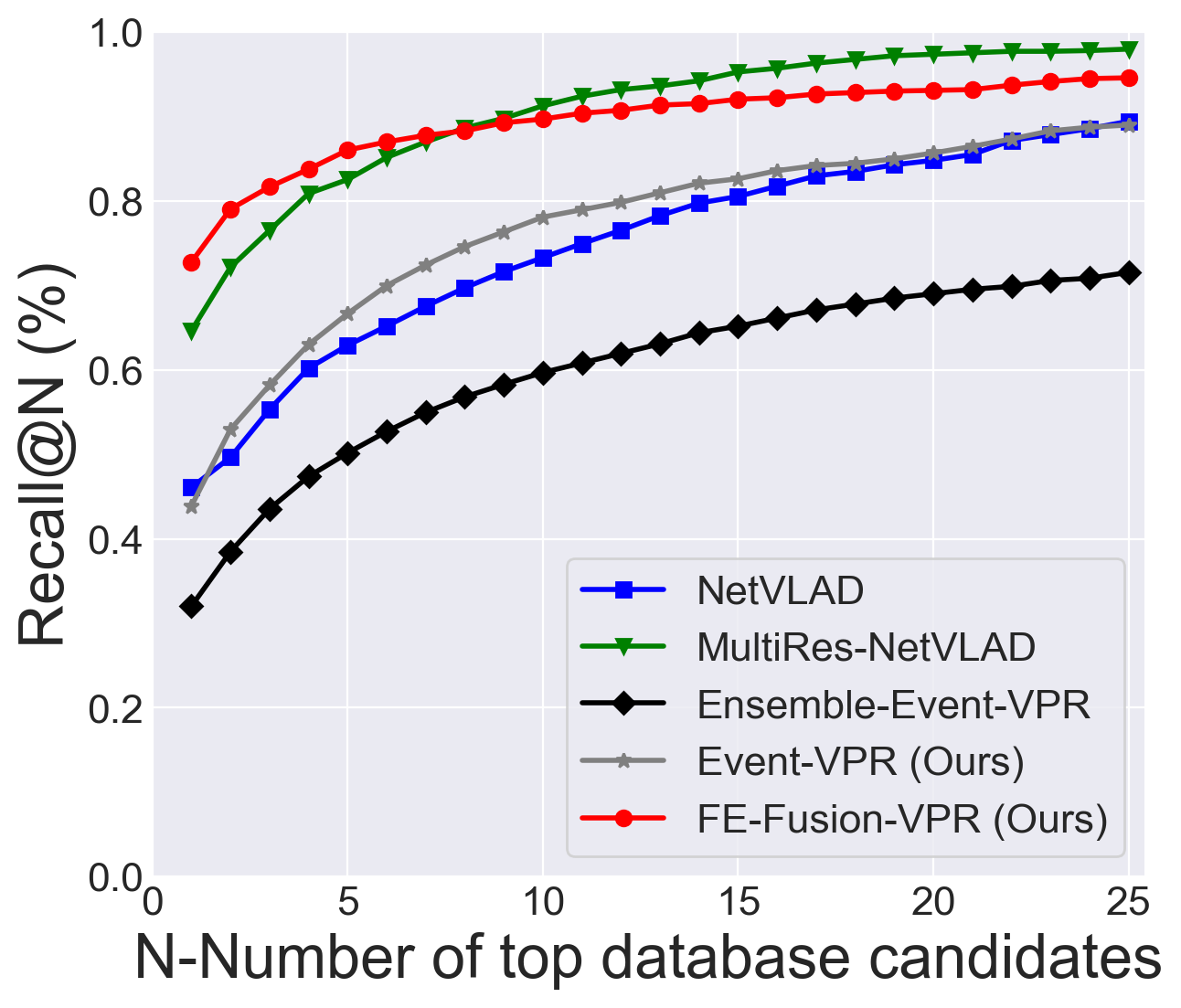}}
\subfigure[**81 / **68]{\label{fig:7f}\includegraphics[width=0.16\textwidth]{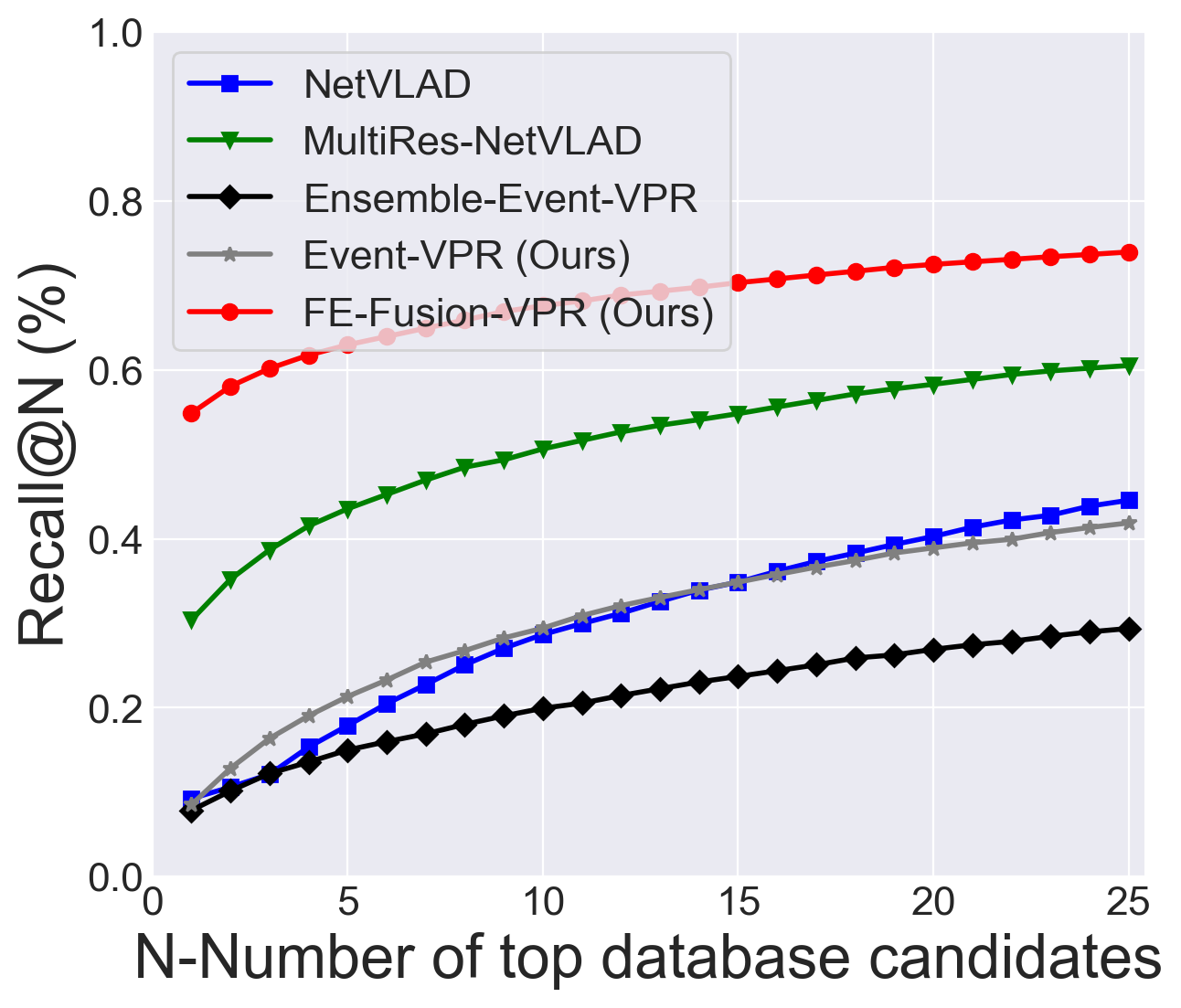}}
\caption{Recall@N curves of NetVLAD \cite{arandjelovic2016netvlad}, MR-NetVLAD \cite{khaliq2022multires}, Ensemble-Event-VPR \cite{fischer2020event}, Event-VPR (ours) \cite{kong2022event} and FE-Fusion-VPR (ours) on Brisbane-Event-VPR \cite{fischer2020event} and DDD20 \cite{hu2020ddd20} datasets. (a)-(d) are on Brisbane-Event-VPR \cite{fischer2020event} dataset, and (e), (f) are on DDD20 \cite{hu2020ddd20} dataset. Our FE-Fusion-VPR (red) is superior in most scenes.}
\label{fig:7}
\vspace{-2mm}
\end{figure*}

\begin{figure*}[htbp]
\vspace{-0mm}
\centering
\subfigure[sunset1 / sunset2]{\label{fig:8a}\includegraphics[width=0.16\textwidth]{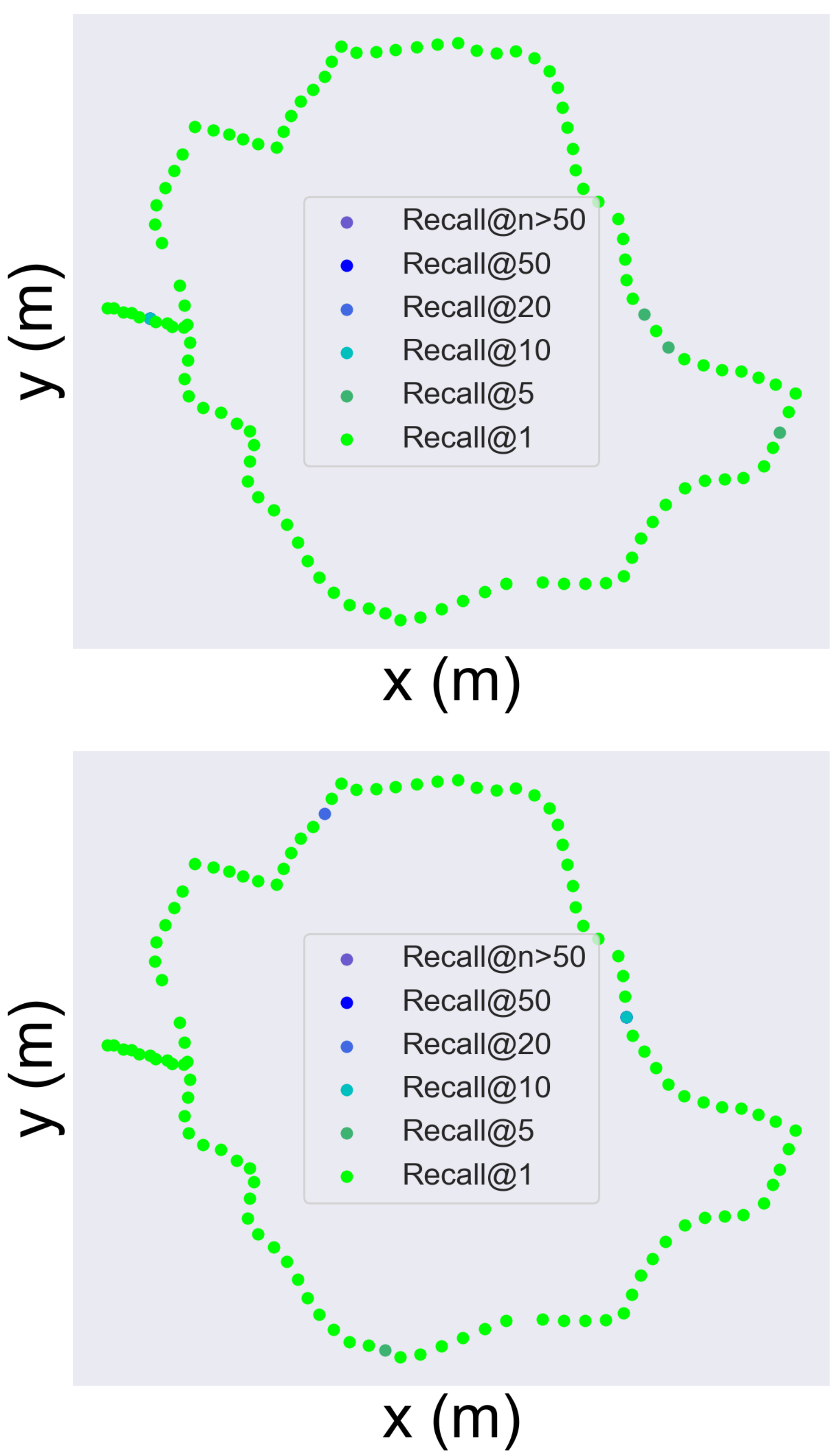}}
\subfigure[sunset1 / sunrise]{\label{fig:8b}\includegraphics[width=0.16\textwidth]{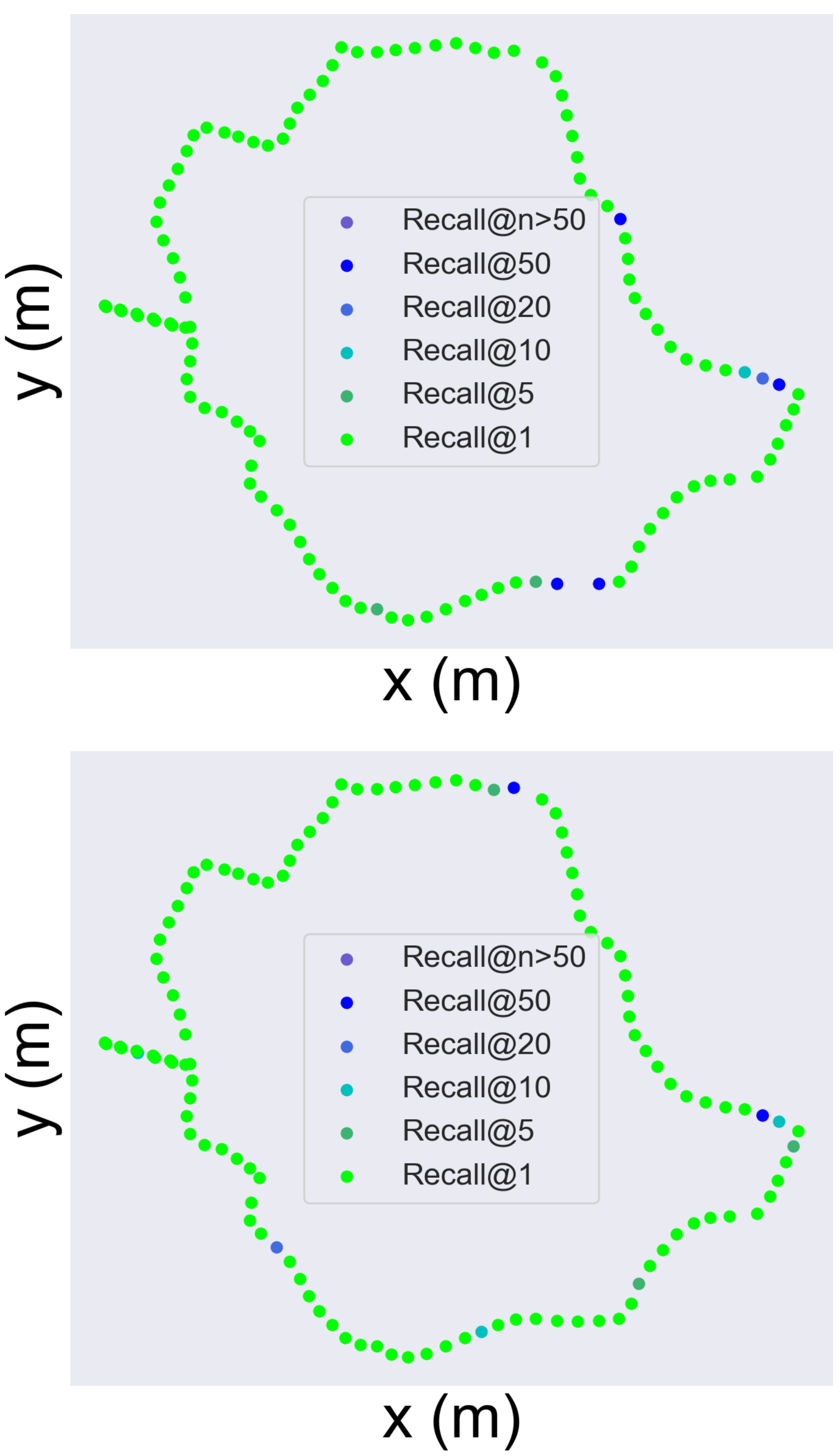}}
\subfigure[sunset1 / morning]{\label{fig:8c}\includegraphics[width=0.16\textwidth]{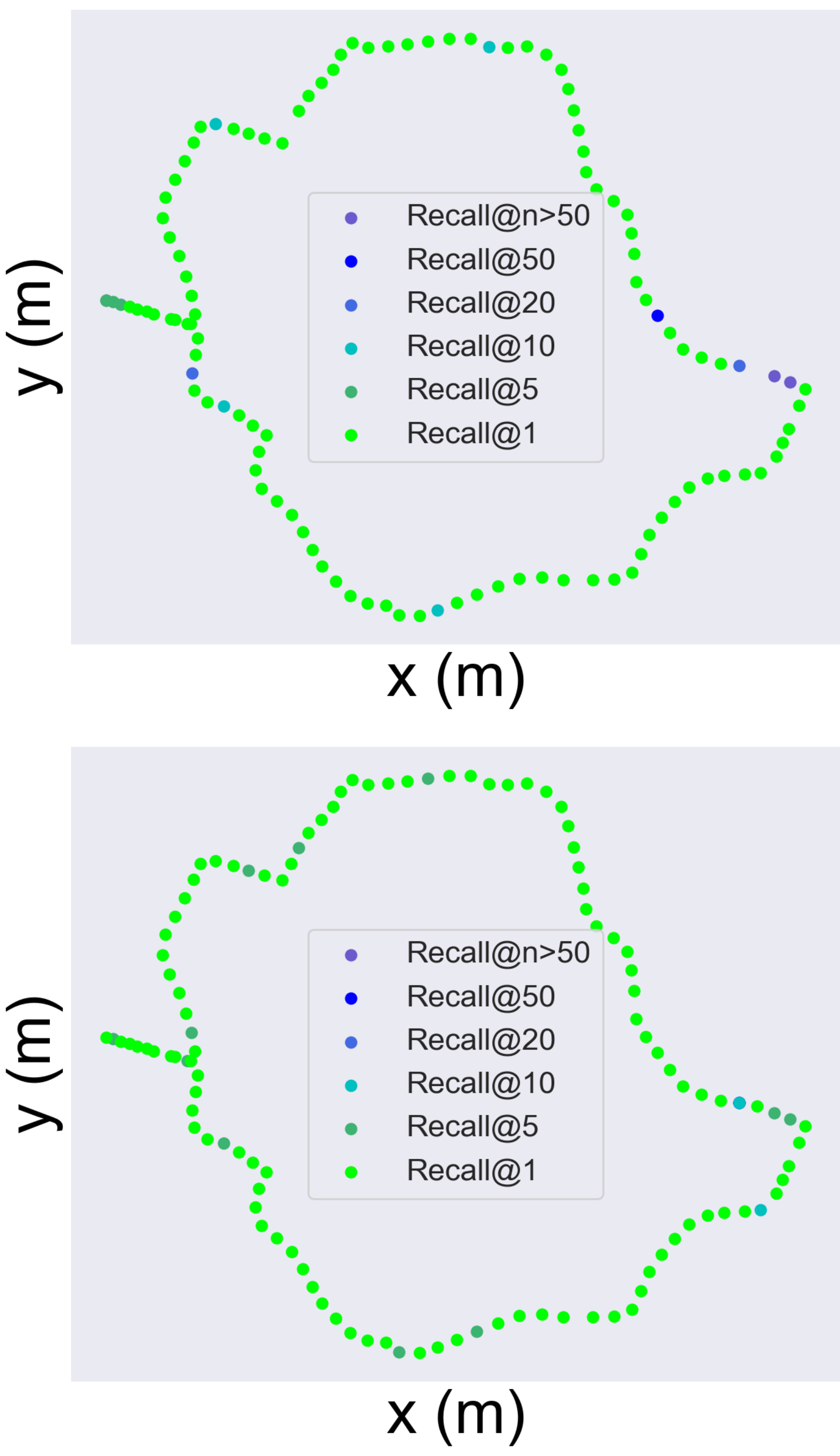}}
\subfigure[sunset1 / daytime]{\label{fig:8d}\includegraphics[width=0.16\textwidth]{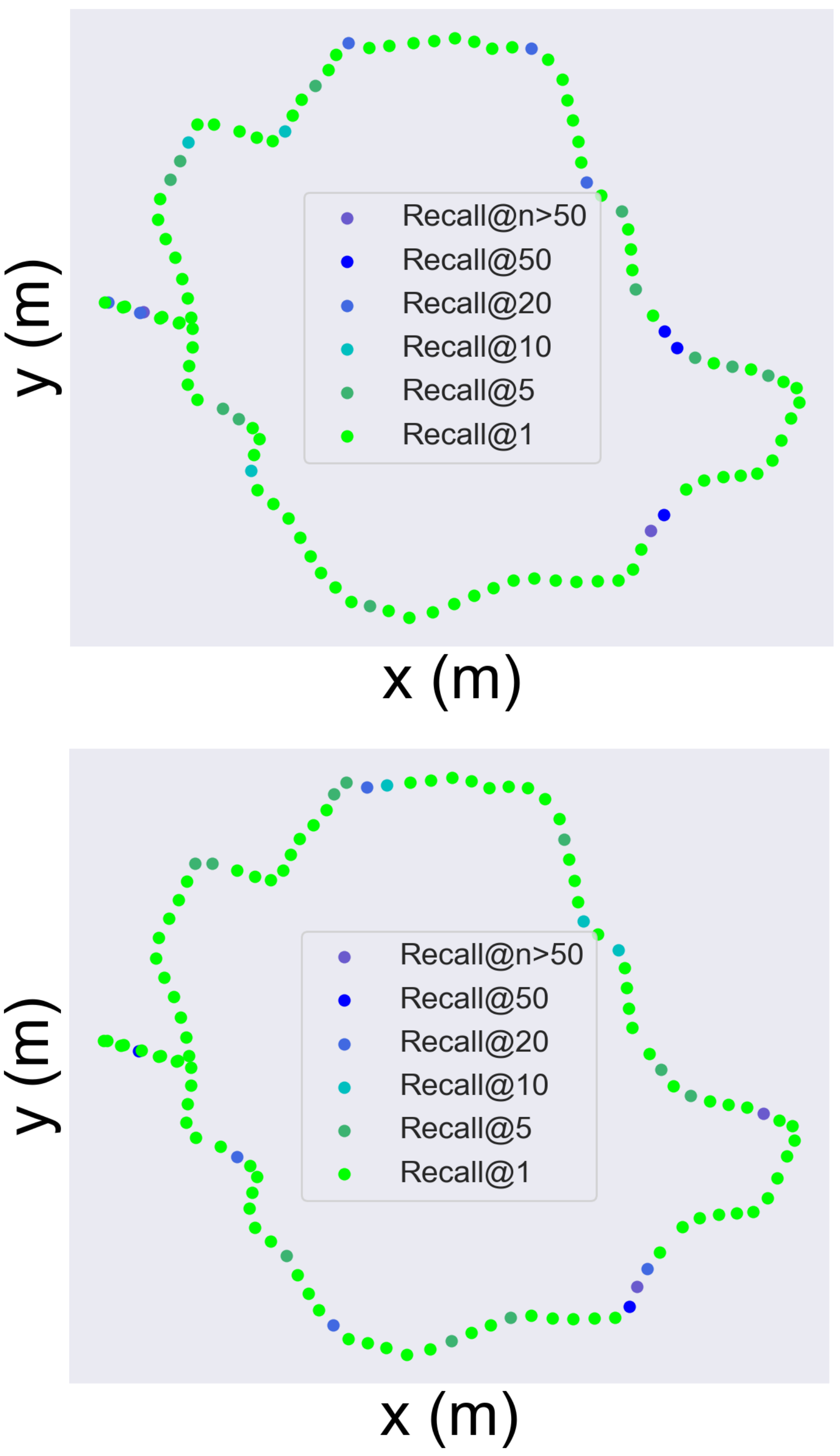}}
\subfigure[**83 / **48]{\label{fig:8e}\includegraphics[width=0.16\textwidth]{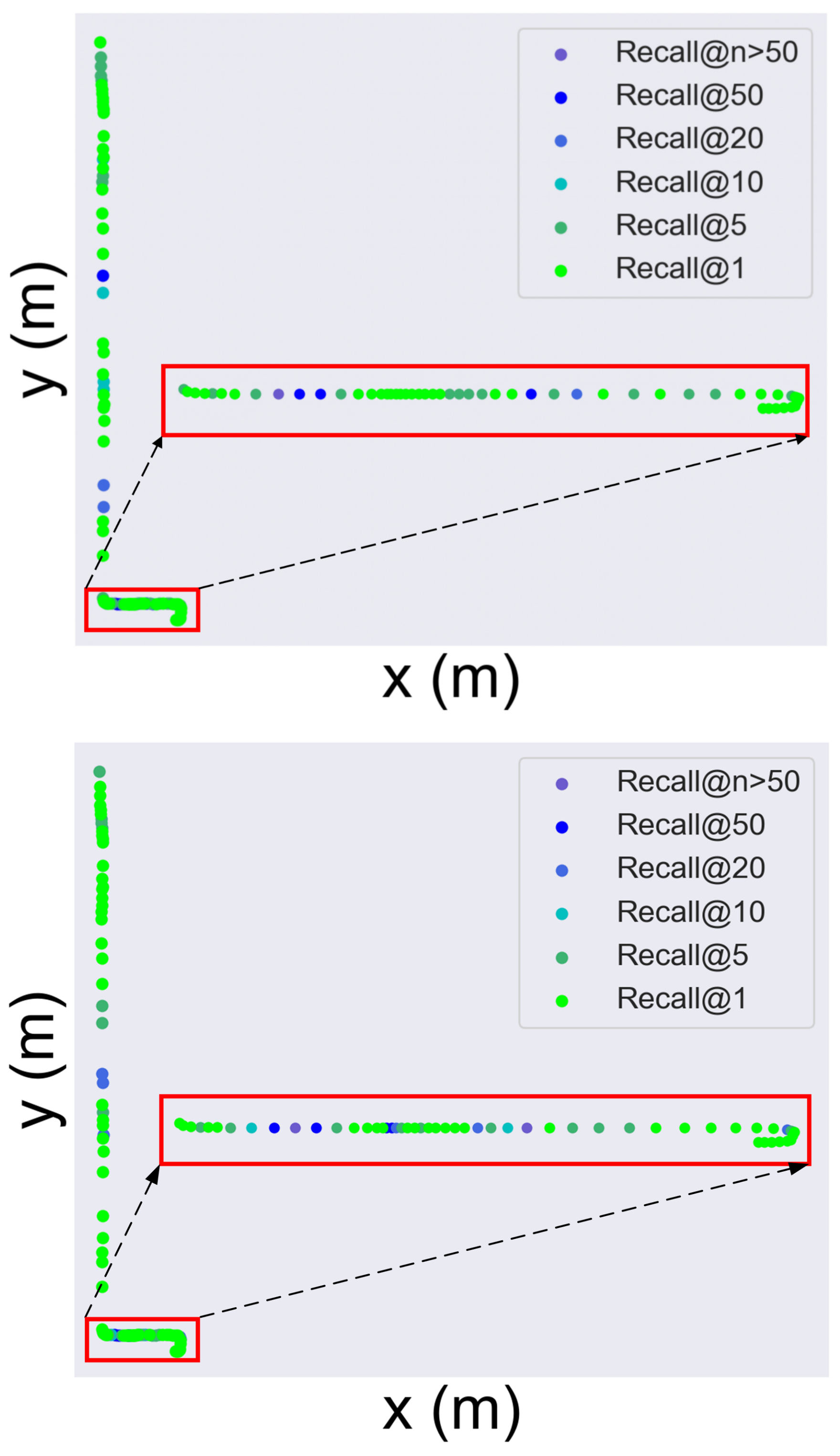}}
\subfigure[**81 / **68]{\label{fig:8f}\includegraphics[width=0.16\textwidth]{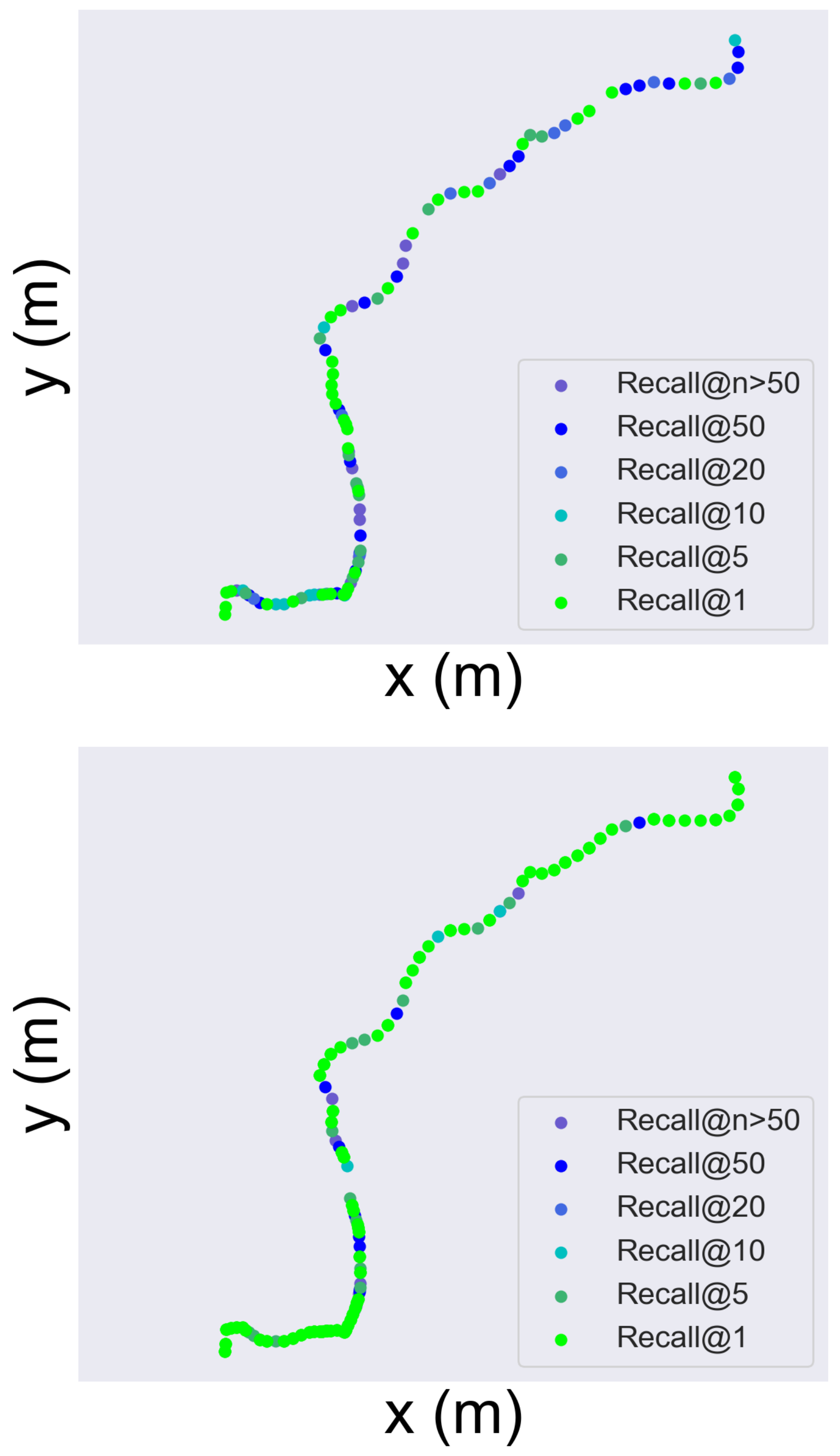}}
\caption{Retrieval success-rate maps of MR-NetVLAD \cite{khaliq2022multires} and FE-Fusion-VPR (ours) on Brisbane-Event-VPR \cite{fischer2020event} and DDD20 \cite{hu2020ddd20} datasets. (a)-(d) are on Brisbane-Event-VPR \cite{fischer2020event} dataset, and (e), (f) are on DDD20 \cite{hu2020ddd20} dataset. Top: MR-NetVLAD \cite{khaliq2022multires}, bottom: FE-Fusion-VPR (ours).}
\label{fig:8}
\vspace{-0mm}
\end{figure*}

\subsection{Comparison against SOTA Methods}
In this section, we present the evaluation details, including frame-based and event-based SOTA VPR methods. And then, we analyze the reasons why our FE-Fusion-VPR has the best performance. Since the results of VEFNet \cite{huang2022vefnet} on the two datasets are not better than the frame-based SOTA method (neither the results in their paper nor the results we verified), we do not compare it with FE-Fusion-VPR.

\subsubsection{Comparison against Frame-based VPR Methods}
We compare our FE-Fusion-VPR against frame-based SOTA algorithms (NetVLAD \cite{arandjelovic2016netvlad} and MR-NetVLAD \cite{khaliq2022multires}), and the experimental results are shown in Tab. \ref{tab:3}, Fig. \ref{fig:6}, Fig. \ref{fig:7} and Fig. \ref{fig:8}. The results show that our method outperforms the above two algorithms in most cases. Tab. \ref{tab:3} shows that the Recall@1 of FE-Fusion-VPR is 3.37\% and 2.55\% higher than NetVLAD and MR-NetVLAD on average on the Brisbane-Event-VPR dataset. On the DDD20 dataset, the advantages of our FE-Fusion-VPR are more obvious. The Recall@1 of FE-Fusion-VPR is 35.72\% and 15.89\% higher than that of the above two algorithms on average. The reason is that there are sequences with obvious differences in intensity appearance in the Brisbane-Event-VPR dataset, and some glare scenarios in the DDD20 dataset, which will limit the performance of frame-based VPR methods. However, event cameras hardly affected by illumination changes can significantly improve the performance of the VPR algorithms. Besides, the DDD20 dataset contains highway scenes with many small objects on both sides of the road. Therefore, our MSF-Net combined with the DRW-Net can improve the performance of VPR due to the multi-scale feature fusion.

\subsubsection{Comparison against Event-based VPR Methods}
As shown in Tab. \ref{tab:3}, Fig. \ref{fig:6}, and Fig. \ref{fig:7}, our FE-Fusion-VPR is much more robust than the pure event-based SOTA methods (Ensemble-Event-VPR \cite{fischer2020event} and Event-VPR \cite{kong2022event}). On the Brisbane-Event-VPR dataset, the Recall@1 of FE-Fusion-VPR is 25.28\% and 28.65\% higher than Event-VPR and Ensemble-Event-VPR on average. On the DDD20 dataset, Recall@1 of FE-Fusion-VPR increases by an average of 37.21\% and 43.48\% over Event-VPR and Ensemble-VPR. The accuracy of our algorithm is much better than the above two methods, illustrating the importance of information from the intensity frame to improve the performance of VPR networks. As it is known to all, event cameras can hardly capture information at low-speed intersections and highways with sparse texture, while standard cameras can capture more background information when the illumination is suitable. Therefore, our FE-Fusion-VPR can achieve higher SOTA VPR performance by combining the advantages of both sensors. Our descriptors can remain rich and vital information by using MSF-Net and DRW-Net.

\begin{table*}[htbp]
\vspace{0mm}
\begin{center}
\caption{Ablation studies on the impact of TSFE-Net, MSF-Net, and DRW-Net on the Performance of FE-Fusion-VPR on Brisbane-Event-VPR \cite{fischer2020event} with the best results \textbf{bolded}. \faTimes: without attention layer, \faCheck: with attention layer.}
\setlength{\tabcolsep}{1pt}
\newcommand{\tabincell}[2]{\begin{tabular}{@{}#1@{}}#2\end{tabular}}
\small
\setlength{\tabcolsep}{1.7mm}{
\begin{tabular}{|c|c|c|c|c|c|p{1cm}}
\hline
\multirow{4}{*}{\tabincell{c}{Ablation \\ Studies}}
&\multirow{4}{*}{\tabincell{c}{Settings}}
&\multicolumn{4}{c|}{\tabincell{c}{Recall@1 (\%)}}  \\
\cline{3-6}
&
&\multicolumn{4}{c|}{\tabincell{c}{Training Set \& Testing Set (database / query)}} \\
\cline{3-6}
&
&\multirow{2}{*}{\tabincell{c}{(dt \& mr) / sr \\ ss1 / ss2}} 
&\multirow{2}{*}{\tabincell{c}{(ss2 \& mr) / sr \\ ss1 / dt}} 
&\multirow{2}{*}{\tabincell{c}{(ss2 \& dt) / sr \\ ss1 / mr}} 
&\multirow{2}{*}{\tabincell{c}{(ss2 \& dt) / mr \\ ss1 / sr}}   \\
&   &   &   &   &   \\
\hline
\multirow{3}{*}{\tabincell{c}{TSFE-Net}}
&Only Frame Encoder &85.63  &78.38  &76.23  &67.98  \\
\cline{2-6}
&Only Event Encoder &93.16  &63.74  &82.16  &53.90  \\
\cline{2-6}
&Frame Encoder + Event Encoder  &\textbf{95.64} &\textbf{93.58} &\textbf{87.41} &\textbf{86.15} \\
\hline
\multirow{4}{*}{\tabincell{c}{MSF-Net}}
&Only $8\mathrm{\times}8$, with Attention Layer (\faTimes/\faCheck)
&86.82 / 91.35 ($\uparrow$)
&76.33 / 83.59 ($\uparrow$)
&85.31 / 80.06 ($\downarrow$)
&72.84 / 68.03 ($\downarrow$)   \\
\cline{2-6}
&Only ${16\mathrm{\times}16}$, with Attention Layer (\faTimes/\faCheck)
&86.03 / 95.53 ($\uparrow$)
&73.25 / 88.13 ($\uparrow$)
&82.81 / 72.72 ($\downarrow$)
&81.33 / 78.67 ($\downarrow$)   \\
\cline{2-6}
&Only ${32\mathrm{\times}32}$, with Attention Layer (\faTimes/\faCheck)
&90.05 / 94.85 ($\uparrow$)
&84.16 / 86.08 ($\uparrow$)
&84.95 / 83.82 ($\downarrow$)
&82.02 / 76.01 ($\downarrow$)   \\
\cline{2-6}
&Multi-Scale, with Attention Layer (\faTimes/\faCheck)
&\textbf{94.06} / \textbf{95.64} ($\uparrow$)
&\textbf{90.89} / \textbf{93.58} ($\uparrow$)
&\textbf{85.85} / \textbf{87.41} ($\uparrow$)
&\textbf{80.09} / \textbf{86.15} ($\uparrow$)   \\
\hline
\multirow{2}{*}{\tabincell{c}{DRW-Net}}
&Concatenation in Length  &91.57  &68.51  &75.30  &73.21  \\
\cline{2-6}
&With Re-Weighting Layer    &\textbf{95.64} &\textbf{93.58} &\textbf{87.41} &\textbf{86.15} \\
\hline
\end{tabular}}
\label{tab:4}
\vspace{0mm}
\end{center}
\end{table*}

\subsection{Ablation Studies}
In this section, we explore the impact of TSFE-Net, MSF-Net, and DRW-Net on the performance of FE-Fusion-VPR.

\subsubsection{Impact of TSFE-Net}
The experimental results in Tab. \ref{tab:4} demonstrate that using a single type of sensor data leads to severe performance degradation on the Brisbane-Event-VPR dataset. Moreover, the SOTA performance of our FE-Fusion-VPR is attributed to using the two vision sensors simultaneously rather than a single type of visual sensor.

\subsubsection{Impact of MSF-Net}
The experimental results in Tab. \ref{tab:4} show that, in most cases, our multi-scale FE-Fusion-VPR can achieve better performance than networks using single-scale features (whatever with/without adding attention layers). Since features at different scales focus on information in different regions, using mid-level or high-level features alone is unreliable. Therefore, multi-scale fusion can achieve higher VPR performance. In addition, by adding attention layers at appropriate locations throughout the whole network, the performance of FE-Fusion-VPR can be further improved.

\subsubsection{Impact of DRW-Net}
In this experiment, we remove the DRW-Net and use the original VLAD layer. Before the features are input into NetVLAD, we flatten the three different scale feature maps $\{\boldsymbol{M}_1,\boldsymbol{M}_2,\boldsymbol{M}_3\}$ (the dimensions are $(\mathrm{D},\mathrm{W}_i,\mathrm{H}_i)$, $i \in \{1,2,3\}$ respectively) output by MSF-Net into three descriptors $\{\boldsymbol{\widetilde{D}}_1,\boldsymbol{\widetilde{D}}_2,\boldsymbol{\widetilde{D}}_3\}$ with dimensions of $(\mathrm{D},\mathrm{m}_i)$, $\mathrm{m}_i=\mathrm{H}_i\times\mathrm{W}_i$, and then we concatenate them along the length dimension:
\begin{equation}
\label{eq:10}
\boldsymbol{\widetilde{D}}=\mathop{\Big{|\Big{|}}}_{i=1}^3\left(\boldsymbol{\widetilde{D}}_i\right)=\mathop{\Big{|\Big{|}}}_{i=1}^3\left(\boldsymbol{f}_\text{flatten}\left(\boldsymbol{M}_i\right)\right),
\end{equation}
where the dimension of the descriptor $\boldsymbol{\widetilde{D}}$ is $(\mathrm{D},\mathrm{m})$, $\mathrm{m}=\mathrm{m}_1+\mathrm{m}_2+\mathrm{m}_3$.
The results in Tab. \ref{tab:4} show that our DRW-Net outperforms methods directly using original VLAD layers for multi-feature fusion in all cases. DRW-Net assigns the weight of each sub-descriptor through auto-learning, which can fully use each descriptor's significant information, so that the final multi-scale descriptor has robust representation ability.

\section{Conclusions}
\label{sec:conclusions}
In this paper, we analyzed the limitation of VPR methods using a frame camera or event camera alone. On that basis, we proposed an attention-based multi-scale network architecture combining frames and events for VPR (named FE-Fusion-VPR) to achieve robust performance in challenging environments. The two key ideas of FE-Fusion-VPR are as follows: First, we achieve visual data fusion based on intensity frames and event volumes. Second, we complete feature fusion based on a multi-scale network and descriptor re-weighting network, which is validated to be effective in our ablation studies.
Compared with existing frame-based and event-based SOTA methods, our FE-Fusion VPR achieves higher performance, especially in scenes with few textures and difficult sunlight glare conditions. In future, we will try to lightweight and accelerate our algorithm for deployment to autonomous vehicles or mini-UAVs. Furthermore, we will also try to realize a deep spiking VPR network architecture \cite{jiang2022neuro} for high energy efficiency inference.
 
\bibliographystyle{IEEEtran}
\bibliography{mybibfile}

\begin{thebibliography}{10}
\providecommand{\url}[1]{#1}
\csname url@samestyle\endcsname
\providecommand{\newblock}{\relax}
\providecommand{\bibinfo}[2]{#2}
\providecommand{\BIBentrySTDinterwordspacing}{\spaceskip=0pt\relax}
\providecommand{\BIBentryALTinterwordstretchfactor}{4}
\providecommand{\BIBentryALTinterwordspacing}{\spaceskip=\fontdimen2\font plus
\BIBentryALTinterwordstretchfactor\fontdimen3\font minus
  \fontdimen4\font\relax}
\providecommand{\BIBforeignlanguage}[2]{{%
\expandafter\ifx\csname l@#1\endcsname\relax
\typeout{** WARNING: IEEEtran.bst: No hyphenation pattern has been}%
\typeout{** loaded for the language `#1'. Using the pattern for}%
\typeout{** the default language instead.}%
\else
\language=\csname l@#1\endcsname
\fi
#2}}
\providecommand{\BIBdecl}{\relax}
\BIBdecl

\bibitem{lowry2015visual}
S.~Lowry, N.~S{\"u}nderhauf, P.~Newman, J.~J. Leonard, D.~Cox, P.~Corke, and
  M.~J. Milford, ``Visual place recognition: A survey,'' \emph{IEEE
  transactions on robotics}, vol.~32, no.~1, pp. 1--19, 2015.

\bibitem{zhang2021visual}
X.~Zhang, L.~Wang, and Y.~Su, ``Visual place recognition: A survey from deep
  learning perspective,'' \emph{Pattern Recognition}, vol. 113, p. 107760,
  2021.

\bibitem{masone2021survey}
C.~Masone and B.~Caputo, ``A survey on deep visual place recognition,''
  \emph{IEEE Access}, vol.~9, pp. 19\,516--19\,547, 2021.

\bibitem{garg2021your}
S.~Garg, T.~Fischer, and M.~Milford, ``Where is your place, visual place
  recognition?'' \emph{arXiv preprint arXiv:2103.06443}, 2021.

\bibitem{delbruck2010activity}
T.~Delbr{\"u}ck, B.~Linares-Barranco, E.~Culurciello, and C.~Posch,
  ``Activity-driven, event-based vision sensors,'' in \emph{IEEE International
  Symposium on Circuits and Systems}, 2010, pp. 2426--2429.

\bibitem{gallego2020event}
G.~Gallego, T.~Delbr{\"u}ck, G.~Orchard, C.~Bartolozzi, B.~Taba, A.~Censi,
  S.~Leutenegger, A.~J. Davison, J.~Conradt, K.~Daniilidis \emph{et~al.},
  ``Event-based vision: A survey,'' \emph{IEEE transactions on pattern analysis
  and machine intelligence}, vol.~44, no.~1, pp. 154--180, 2020.

\bibitem{chen2020event}
G.~Chen, H.~Cao, J.~Conradt, H.~Tang, F.~Rohrbein, and A.~Knoll, ``Event-based
  neuromorphic vision for autonomous driving: A paradigm shift for bio-inspired
  visual sensing and perception,'' \emph{IEEE Signal Processing Magazine},
  vol.~37, no.~4, pp. 34--49, 2020.

\bibitem{wu2021novel}
T.-H. Wu, C.~Gong, D.~Kong, S.~Xu, and Q.~Liu, ``A novel visual object
  detection and distance estimation method for hdr scenes based on event
  camera,'' in \emph{International Conference on Computer and Communications},
  2021, pp. 636--640.

\bibitem{gehrig2018asynchronous}
D.~Gehrig, H.~Rebecq, G.~Gallego, and D.~Scaramuzza, ``Asynchronous,
  photometric feature tracking using events and frames,'' in \emph{European
  Conference on Computer Vision}, 2018, pp. 750--765.

\bibitem{jiang2019mixed}
Z.~Jiang, P.~Xia, K.~Huang, W.~Stechele, G.~Chen, Z.~Bing, and A.~Knoll,
  ``Mixed frame-/event-driven fast pedestrian detection,'' in
  \emph{International Conference on Robotics and Automation}, 2019, pp.
  8332--8338.

\bibitem{hu2020ddd20}
Y.~Hu, J.~Binas, D.~Neil, S.-C. Liu, and T.~Delbruck, ``Ddd20 end-to-end event
  camera driving dataset: Fusing frames and events with deep learning for
  improved steering prediction,'' in \emph{International Conference on
  Intelligent Transportation Systems}, 2020, pp. 1--6.

\bibitem{gehrig2021combining}
D.~Gehrig, M.~R{\"u}egg, M.~Gehrig, J.~Hidalgo-Carri{\'o}, and D.~Scaramuzza,
  ``Combining events and frames using recurrent asynchronous multimodal
  networks for monocular depth prediction,'' \emph{IEEE Robotics and Automation
  Letters}, vol.~6, no.~2, pp. 2822--2829, 2021.

\bibitem{pan2020single}
L.~Pan, M.~Liu, and R.~Hartley, ``Single image optical flow estimation with an
  event camera,'' in \emph{IEEE/CVF Conference on Computer Vision and Pattern
  Recognition}, 2020, pp. 1669--1678.

\bibitem{lee2022fusion}
C.~Lee, A.~K. Kosta, and K.~Roy, ``Fusion-flownet: Energy-efficient optical
  flow estimation using sensor fusion and deep fused spiking-analog network
  architectures,'' in \emph{International Conference on Robotics and
  Automation}, 2022, pp. 6504--6510.

\bibitem{vidal2018ultimate}
A.~R. Vidal, H.~Rebecq, T.~Horstschaefer, and D.~Scaramuzza, ``Ultimate slam?
  combining events, images, and imu for robust visual slam in hdr and
  high-speed scenarios,'' \emph{IEEE Robotics and Automation Letters}, vol.~3,
  no.~2, pp. 994--1001, 2018.

\bibitem{jung2020constrained}
J.~H. Jung and C.~G. Park, ``Constrained filtering-based fusion of images,
  events, and inertial measurements for pose estimation,'' in \emph{IEEE
  International Conference on Robotics and Automation}, 2020, pp. 644--650.

\bibitem{lowe2004distinctive}
D.~G. Lowe, ``Distinctive image features from scale-invariant keypoints,''
  \emph{International journal of computer vision}, vol.~60, no.~2, pp. 91--110,
  2004.

\bibitem{bay2006surf}
H.~Bay, T.~Tuytelaars, and L.~V. Gool, ``Surf: Speeded up robust features,'' in
  \emph{European conference on computer vision}, 2006, pp. 404--417.

\bibitem{rublee2011orb}
E.~Rublee, V.~Rabaud, K.~Konolige, and G.~Bradski, ``Orb: An efficient
  alternative to sift or surf,'' in \emph{International conference on computer
  vision}, 2011, pp. 2564--2571.

\bibitem{dalal2005histograms}
N.~Dalal and B.~Triggs, ``Histograms of oriented gradients for human
  detection,'' in \emph{IEEE computer society conference on computer vision and
  pattern recognition}, vol.~1, 2005, pp. 886--893.

\bibitem{oliva2006building}
A.~Oliva and A.~Torralba, ``Building the gist of a scene: The role of global
  image features in recognition,'' \emph{Progress in brain research}, vol. 155,
  pp. 23--36, 2006.

\bibitem{angeli2008fast}
A.~Angeli, D.~Filliat, S.~Doncieux, and J.-A. Meyer, ``Fast and incremental
  method for loop-closure detection using bags of visual words,'' \emph{IEEE
  transactions on robotics}, vol.~24, no.~5, pp. 1027--1037, 2008.

\bibitem{galvez2012bags}
D.~G{\'a}lvez-L{\'o}pez and J.~D. Tardos, ``Bags of binary words for fast place
  recognition in image sequences,'' \emph{IEEE Transactions on Robotics},
  vol.~28, no.~5, pp. 1188--1197, 2012.

\bibitem{perronnin2010large}
F.~Perronnin, Y.~Liu, J.~S{\'a}nchez, and H.~Poirier, ``Large-scale image
  retrieval with compressed fisher vectors,'' in \emph{IEEE computer society
  conference on computer vision and pattern recognition}, 2010, pp. 3384--3391.

\bibitem{sanchez2013image}
J.~S{\'a}nchez, F.~Perronnin, T.~Mensink, and J.~Verbeek, ``Image
  classification with the fisher vector: Theory and practice,''
  \emph{International journal of computer vision}, vol. 105, no.~3, pp.
  222--245, 2013.

\bibitem{jegou2010aggregating}
H.~J{\'e}gou, M.~Douze, C.~Schmid, and P.~P{\'e}rez, ``Aggregating local
  descriptors into a compact image representation,'' in \emph{IEEE computer
  society conference on computer vision and pattern recognition}, 2010, pp.
  3304--3311.

\bibitem{arandjelovic2013all}
R.~Arandjelovic and A.~Zisserman, ``All about vlad,'' in \emph{IEEE conference
  on Computer Vision and Pattern Recognition}, 2013, pp. 1578--1585.

\bibitem{torii201524}
A.~Torii, R.~Arandjelovic, J.~Sivic, M.~Okutomi, and T.~Pajdla, ``24/7 place
  recognition by view synthesis,'' in \emph{IEEE conference on computer vision
  and pattern recognition}, 2015, pp. 1808--1817.

\bibitem{khaliq2019holistic}
A.~Khaliq, S.~Ehsan, Z.~Chen, M.~Milford, and K.~McDonald-Maier, ``A holistic
  visual place recognition approach using lightweight cnns for significant
  viewpoint and appearance changes,'' \emph{IEEE transactions on robotics},
  vol.~36, no.~2, pp. 561--569, 2019.

\bibitem{arandjelovic2016netvlad}
R.~Arandjelovic, P.~Gronat, A.~Torii, T.~Pajdla, and J.~Sivic, ``Netvlad: Cnn
  architecture for weakly supervised place recognition,'' in \emph{IEEE
  conference on computer vision and pattern recognition}, 2016, pp. 5297--5307.

\bibitem{hausler2021patch}
S.~Hausler, S.~Garg, M.~Xu, M.~Milford, and T.~Fischer, ``Patch-netvlad:
  Multi-scale fusion of locally-global descriptors for place recognition,'' in
  \emph{IEEE/CVF Conference on Computer Vision and Pattern Recognition}, 2021,
  pp. 14\,141--14\,152.

\bibitem{khaliq2022multires}
A.~Khaliq, M.~Milford, and S.~Garg, ``Multires-netvlad: Augmenting place
  recognition training with low-resolution imagery,'' \emph{IEEE Robotics and
  Automation Letters}, vol.~7, no.~2, pp. 3882--3889, 2022.

\bibitem{fischer2020event}
T.~Fischer and M.~Milford, ``Event-based visual place recognition with
  ensembles of temporal windows,'' \emph{IEEE Robotics and Automation Letters},
  vol.~5, no.~4, pp. 6924--6931, 2020.

\bibitem{lee2021eventvlad}
A.~J. Lee and A.~Kim, ``Eventvlad: Visual place recognition with reconstructed
  edges from event cameras,'' in \emph{IEEE/RSJ International Conference on
  Intelligent Robots and Systems}, 2021, pp. 2247--2252.

\bibitem{kong2022event}
D.~Kong, Z.~Fang, K.~Hou, H.~Li, J.~Jiang, S.~Coleman, and D.~Kerr,
  ``Event-vpr: End-to-end weakly supervised deep network architecture for
  visual place recognition using event-based vision sensor,'' \emph{IEEE
  Transactions on Instrumentation and Measurement}, vol.~71, pp. 1--18, 2022.

\bibitem{huang2022vefnet}
Z.~Huang, R.~Huang, L.~Sun, C.~Zhao, M.~Huang, and S.~Su, ``Vefnet: an
  event-rgb cross modality fusion network for visual place recognition,'' in
  \emph{IEEE International Conference on Image Processing}, 2022, pp.
  2671--2675.

\bibitem{he2016deep}
K.~He, X.~Zhang, S.~Ren, and J.~Sun, ``Deep residual learning for image
  recognition,'' in \emph{IEEE conference on computer vision and pattern
  recognition}, 2016, pp. 770--778.

\bibitem{woo2018cbam}
S.~Woo, J.~Park, J.-Y. Lee, and I.~S. Kweon, ``Cbam: Convolutional block
  attention module,'' in \emph{European conference on computer vision}, 2018,
  pp. 3--19.

\bibitem{maqueda2018event}
A.~I. Maqueda, A.~Loquercio, G.~Gallego, N.~Garc{\'\i}a, and D.~Scaramuzza,
  ``Event-based vision meets deep learning on steering prediction for
  self-driving cars,'' in \emph{IEEE conference on computer vision and pattern
  recognition}, 2018, pp. 5419--5427.

\bibitem{ye2018hierarchical}
M.~Ye, X.~Lan, J.~Li, and P.~Yuen, ``Hierarchical discriminative learning for
  visible thermal person re-identification,'' in \emph{AAAI Conference on
  Artificial Intelligence}, vol.~32, no.~1, 2018.

\bibitem{chen2014convolutional}
Z.~Chen, O.~Lam, A.~Jacobson, and M.~Milford, ``Convolutional neural
  network-based place recognition,'' \emph{arXiv preprint arXiv:1411.1509},
  2014.

\bibitem{sunderhauf2015performance}
N.~S{\"u}nderhauf, S.~Shirazi, F.~Dayoub, B.~Upcroft, and M.~Milford, ``On the
  performance of convnet features for place recognition,'' in \emph{IEEE/RSJ
  international conference on intelligent robots and systems}, 2015, pp.
  4297--4304.

\bibitem{lin2017feature}
T.-Y. Lin, P.~Doll{\'a}r, R.~Girshick, K.~He, B.~Hariharan, and S.~Belongie,
  ``Feature pyramid networks for object detection,'' in \emph{IEEE conference
  on computer vision and pattern recognition}, 2017, pp. 2117--2125.

\bibitem{jin2022you}
Z.~Jin, D.~Yu, L.~Song, Z.~Yuan, and L.~Yu, ``You should look at all objects,''
  in \emph{European Conference on Computer Vision}, 2022, pp. 332--349.

\bibitem{jiang2022neuro}
J.~Jiang, D.~Kong, K.~Hou, X.~Huang, H.~Zhuang, and F.~Zheng, ``Neuro-planner:
  A 3d visual navigation method for mav with depth camera based on neuromorphic
  reinforcement learning,'' \emph{arXiv preprint arXiv:2210.02305}, 2022.

\end{thebibliography}

\end{document}